\documentclass{article}

\usepackage[preprint]{neurips_2025}

\usepackage[utf8]{inputenc} 
\usepackage[T1]{fontenc}    
\usepackage{hyperref}       
\usepackage{url}            
\usepackage{booktabs}       
\usepackage{amsfonts}       
\usepackage{nicefrac}       
\usepackage{microtype}      
\usepackage{xcolor}         
\usepackage{graphicx}
\usepackage{subcaption}
\usepackage{subfloat}
\usepackage[inline]{enumitem}
\usepackage{amsmath}
\usepackage[breakable]{tcolorbox}
\usepackage{changepage}
\usepackage{wrapfig}
\usepackage{pgfplots}
\usepackage{tikz}
\usepackage{enumitem}
\usepackage{kotex}
\usepackage[frozencache,cachedir=minted-0-main]{minted}

\setitemize{leftmargin=*}

\usepackage{tikz}
\usetikzlibrary{positioning, shapes, arrows, fit, backgrounds, calc, shadows, decorations.pathreplacing}
\usepackage{tocloft}

\newcommand{\thearena}{Othello AI Arena}
\newcommand{\meta}{meta-level intelligence}
\newcommand{\task}{task-level strategy}

\title{The Othello AI Arena: Evaluating Intelligent Systems Through Limited-Time Adaptation to Unseen Boards}

\newcommand{\todo}[1]{\textcolor{red}{#1}}

\author{
Sundong Kim \\
   Gwangju Institute of Science and Technology \\
   \texttt{sundong@gist.ac.kr}
}

\begin{document}

\maketitle

\begin{abstract}
\label{Sec:Abstract}
The ability to rapidly adapt to novel and unforeseen environmental changes is a cornerstone of artificial general intelligence (AGI), yet it remains a critical blind spot in most existing AI benchmarks. Traditional evaluation largely focuses on optimizing performance within fixed environments, failing to assess systems' flexibility and generalization capabilities when faced with even subtle rule or structural modifications. Addressing this gap, I introduce the Othello AI Arena, a novel benchmark framework designed to evaluate intelligent systems based on their capacity for limited-time adaptation to unseen environments. Our platform poses a meta-learning challenge: participants must develop systems that can analyze the specific configuration and rules of a novel Othello board within a strict time limit ($\approx 60$ seconds) and generate a tailored, high-performing strategy for that unique environment. With this, evaluation of the meta-level intelligence (which performs the analysis and strategy generation) can be separated from the task-level strategy performance of the generated strategy. The Arena features a diverse set of game stages, including public stages for development and private stages with structural and rule variations (such as altered board sizes, blocked cells, non-standard capture mechanics, and modified turn dynamics) designed to test genuine adaptive and generalization capabilities. Implemented as an accessible web-based platform, the Arena provides real-time visualization, automated evaluation using multi-dimensional metrics (including adaptation speed and efficiency), and comprehensive logging for post-hoc analysis. Initial observations from pilot tests and preliminary student engagements highlight fascinating patterns in adaptation approaches, ranging from rapid parameter tuning to rudimentary environmental model learning through simulation. The Othello AI Arena offers a unique educational tool and a valuable research benchmark for fostering and evaluating the crucial skill of rapid, intelligent adaptation in AI systems.

\end{abstract}

\addtocontents{toc}{\protect\setcounter{tocdepth}{0}} 
\addtocontents{toc}{\protect\makeatletter}
\addtocontents{toc}{\protect\global\protect\@empty\protect\let\protect\@starttocsection\protect\relax}
\addtocontents{toc}{\protect\global\protect\@empty\protect\let\protect\@starttocsubsection\protect\relax}
\addtocontents{toc}{\protect\global\protect\@empty\protect\let\protect\@starttocsubsubsection\protect\relax}
\addtocontents{toc}{\protect\global\protect\@empty\protect\let\protect\@starttocparagraph\protect\relax}
\addtocontents{toc}{\protect\global\protect\@empty\protect\let\protect\@starttocsubparagraph\protect\relax}
\addtocontents{toc}{\protect\makeatother}

\addtocontents{toc}{\protect\setcounter{tocdepth}{2}} 

\setlength{\cftbeforesecskip}{0.5\baselineskip}   
\setlength{\cftbeforesubsecskip}{0.4\baselineskip} 
\setlength{\cftaftertoctitleskip}{0.6\baselineskip} 

\begin{small} 
\renewcommand{\contentsname}{Contents} 
\linespread{0.8} 
\tableofcontents 
\linespread{1.0} 
\end{small}

\section{Introduction}
\label{sec:introduction}

Current AI benchmarks predominantly focus on measuring peak performance in static environments. While valuable, this approach overlooks the dynamic nature of real-world problems, where environments are constantly evolving, requiring intelligent agents to be robust and adaptable. Furthermore, many benchmarks evaluate only the final output or performance, without explicitly assessing the underlying process by which an intelligent system analyzes a new task and generates a suitable strategy. This makes it difficult to understand and evaluate the meta-cognitive abilities, such as efficient knowledge transfer, rapid model building, and strategic reasoning under uncertainty, that are critical for true general intelligence.

From an educational standpoint, providing students with opportunities to design systems that can adapt to changing circumstances offers a richer learning experience than simply optimizing for fixed problems. Such challenges cultivate not just problem-solving skills but also meta-reasoning~\cite{finn2017maml}, time management under computational constraints, and a deeper understanding of the exploration-exploitation trade-off. Incorporating adaptive challenges is particularly relevant in AGI education, where fostering the ability to handle novelty is a core objective.

Addressing these limitations, I introduce the \thearena, a novel benchmark framework specifically designed to evaluate the adaptive intelligence of AI systems. This is grounded in the belief that a key aspect of AGI is the ability to quickly analyze an unfamiliar environment and synthesize an effective strategy within a limited timeframe. The \thearena{} provides a platform where participants must develop intelligent systems capable of this rapid adaptation. The Othello AI Arena is accessible at \url{https://sundong.kim/courses/agi25sp/othello-leaderboard/hw2.html}

\begin{figure}[h] 
\centering 
\includegraphics[width=\textwidth]{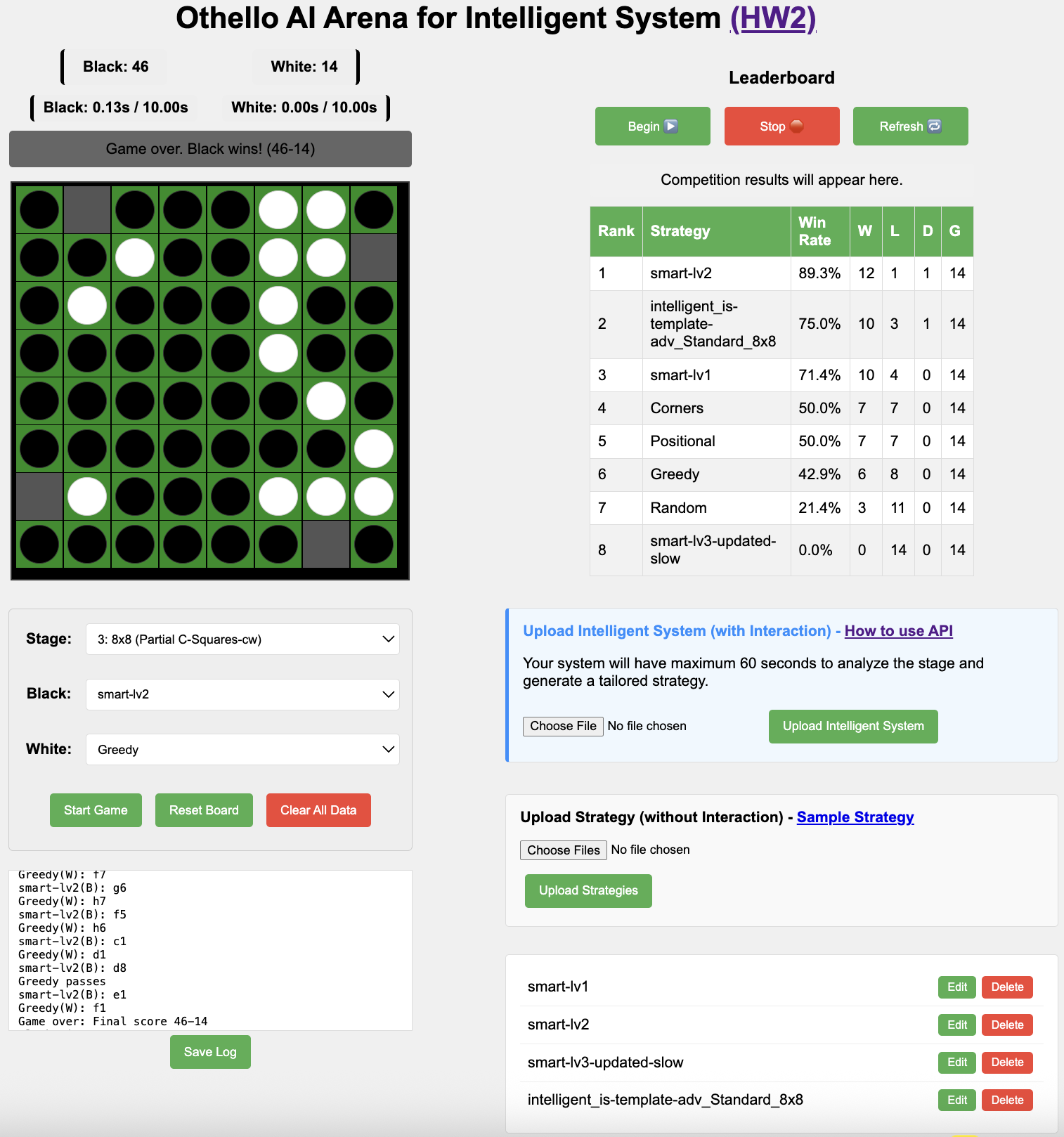} 
\caption{Screenshot of the \thearena{} web interface on the "$8\times 8$ (Partial C-Squares-cw)" stage. The leaderboard displays the performance of various strategies, including a generated intelligent system strategy.} 
\label{fig:arena_screenshot} 
\end{figure}

\section{The \thearena{} Framework}
\label{sec:framework}

The \thearena{} is a comprehensive platform for evaluating intelligent systems' adaptive capabilities in Othello~\cite{rose2005othello}. It comprises core components, a structured challenge design with diverse environmental variations, and a defined API for system-environment interaction.

\subsection{System Architecture}
\label{sub:architecture}

The modular architecture includes the User Interface (UI), Core Game Engine, Intelligent System Execution Environment, Evaluation Module, and Logging and Analysis System. Figure~\ref{fig:architecture} illustrates their interaction. The UI (\texttt{hw2.html}, \texttt{app.js}, \texttt{game-ui.js}) provides web-based visualization and management. The Core Game Engine (\texttt{game-core.js}) implements Othello rules and stage variations. The Execution Environment (\texttt{intelligent-system-loader.js}) safely runs submitted code, enforcing $T_{\text{analysis}}$ ($\approx 60$ seconds) for strategy generation (\texttt{analyzeStage}) and $T_{\text{game}}$ ($\approx 10$ seconds total) for game execution. The Evaluation Module (\texttt{tournament.js}) manages tournaments and metrics. The Logging and Analysis System (\texttt{game-logger.js}) records detailed game data for post-hoc analysis (\texttt{Using-game-logs-for-world-model-learning.md}). The system flow involves system upload, \texttt{analyzeStage} execution per stage within $T_{\text{analysis}}$, returning a strategy, and game execution using the strategy under $T_{\text{game}}$, with logging throughout.

\begin{figure}[ht]
\centering
\includegraphics[width=\textwidth]{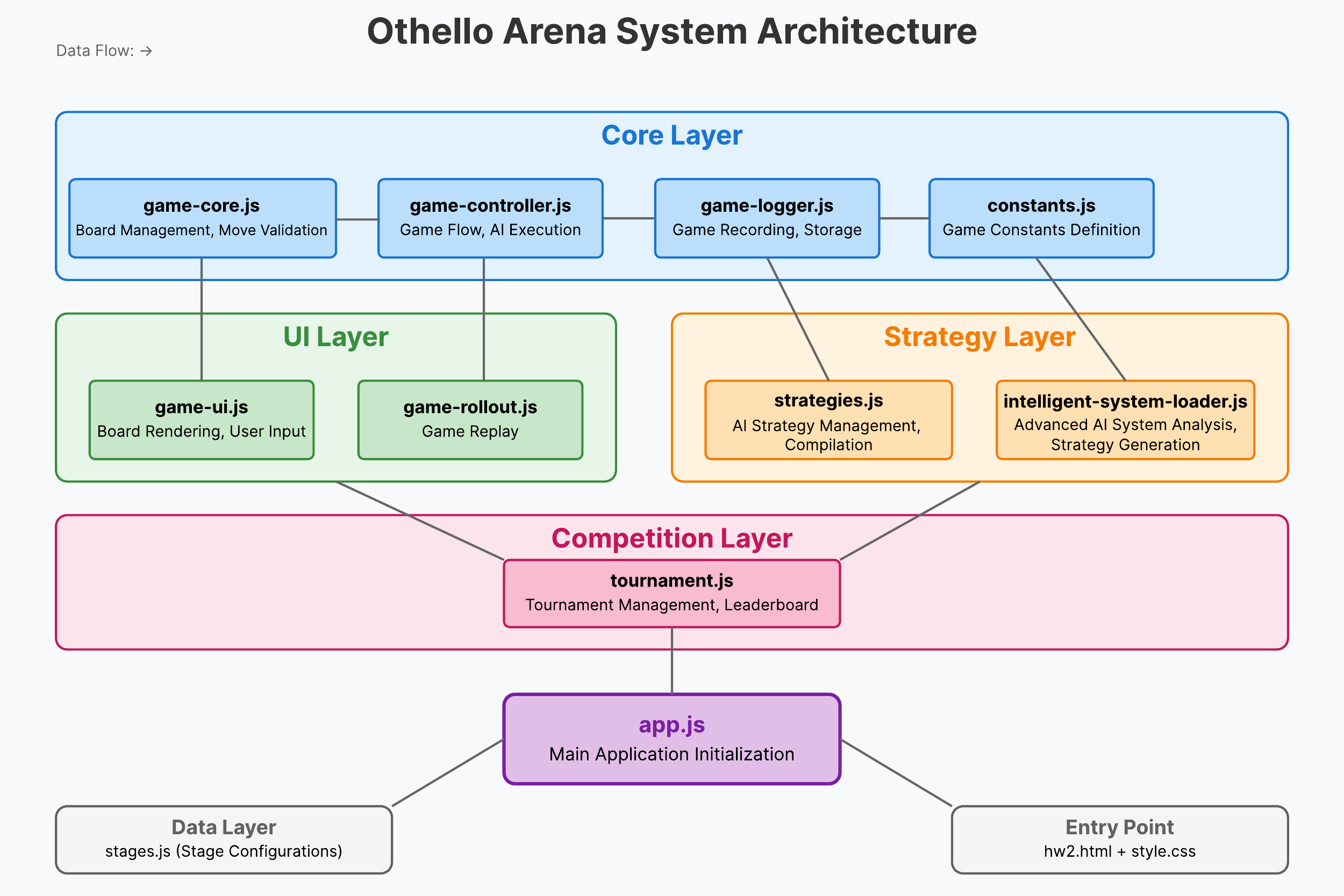} 
\caption{Overall system architecture of the Othello AI Arena, depicting modular layers and data flow. Intelligent systems interact primarily with components within the Strategy Layer.}
\label{fig:othello-arch}
\end{figure}

\subsection{Challenge Design}
\label{sub:challenge_design}

The challenge uses public (for development) and private (unseen for evaluation) stages to assess real-time adaptation. Stages feature variations probing different cognitive aspects. I categorize variations (\texttt{stages-extended.js}) not limited to below settings:

\begin{itemize}
    \item \textbf{Structural Variations:} Modifications to the physical layout of the board.
    \begin{itemize}
        \item \textit{Board Size:} Different grid dimensions (e.g., $6\times 6$, $10\times 10$ from standard $8\times 8$). This tests spatial generalization and the ability to adapt positional evaluation and search strategies to new scales.
        \item \textit{Blocked Cells:} Introducing impassable cells on the board. This challenges pathfinding, region control, and the re-evaluation of move validity and positional value in a constrained space.
    \end{itemize}
    \item \textbf{Rule Variations:} Alterations to the fundamental game mechanics.
    \begin{itemize}
        \item \textit{Capture Mechanics:} Modifying how pieces are captured (e.g., `ignore occlusion' where blocked cells don't stop capture lines). This demands rapid rule induction and adaptation of move simulation and evaluation logic.
        \item \textit{Turn Dynamics:} Changes to whose turn it is (e.g., `fewer pieces continue' where the player with fewer pieces takes consecutive turns). This affects temporal planning and requires understanding and predicting the flow of the game based on piece counts.
        \item \textit{Winning Conditions:} Altering the goal of the game (e.g., Reverse Othello, where the player with the \textit{least} pieces wins). This challenges goal reorientation and counter-intuitive strategic thinking.
    \end{itemize}
    \item \textbf{Initial State Variations:} Non-standard starting configurations.
     \begin{itemize}
        \item \textit{Non-standard Initial Placement:} Pieces starting in unusual positions. This invalidates standard opening books and requires dynamic early-game strategy generation.
        \item \textit{Pre-placed Special Pieces:} Introducing pieces with unique properties or interactions (potential future extension).
    \end{itemize}
\end{itemize}
Each variation imposes a cognitive hurdle for the intelligent system within $T_{\text{analysis}}$. The generated \task{} function takes board state, player, and valid moves, returning a move within the cumulative $T_{\text{game}}$ limit.

\subsection{API and Environment Interaction}
\label{sub:api_interaction}

The \thearena{} provides a well-defined API\footnote{\scriptsize \url{https://sundong.kim/courses/agi25sp/othello-leaderboard/Intelligent-system-API-reference/}}\footnote{\scriptsize \url{https://sundong.kim/courses/agi25sp/othello-leaderboard/Using-game-logs-for-world-model-learning/}} that serves as the sole channel through which the intelligent system can interact with and learn about a new stage during the 60-second analysis ($T_{\text{analysis}}$) phase. Key functions include:

\begin{itemize}
    \item \texttt{getValidMoves(board, player)}: Returns a list of valid moves for a player on a given board state, respecting the stage's specific rules.
    \item \texttt{simulateMove(board, player, row, col)}: Simulates the outcome of making a specific move, returning the resulting board state and number of pieces captured. This is crucial for exploring game dynamics and evaluating potential move consequences without playing the game.
    \item \texttt{evaluateBoard(board, player)}: Provides an evaluation of a given board state using pre-defined metrics (e.g., piece count, mobility, corner control). Note that the weights or interpretation of these metrics might need to be learned or adjusted by the intelligent system based on the stage's characteristics (as seen in \texttt{intelligent-system-template-adv.js}).
\end{itemize}

Systems effectively use these APIs within $T_{\text{analysis}}$ (e.g., via simulation-based learning as in \texttt{intelligent-system-template-adv.js}) to model the environment and synthesize a strategy. A successful intelligent system effectively utilizes these tools within the time limit. For instance, the template in \texttt{intelligent-system-template-adv.js} uses \texttt{api.simulateMove} and \texttt{api.getValidMoves} extensively in its self-play loop to explore the stage's game dynamics and collect data (e.g., position win rates, evidence for rule variations) from which it could learn a tailored strategy. This process exemplifies how systems must learn an environmental model and synthesize a strategy through active interaction constrained by time and the API.

Game logging (\texttt{game-logger.js}) captures detailed information about every move and board state during gameplay. While not directly part of the `analyzeStage` API for the \emph{current} stage analysis, these logs from \emph{previous} games on public stages can potentially be leveraged by an intelligent system during its analysis phase to transfer knowledge or identify patterns (\texttt{Using-game-logs-for-world-model-learning.md}).

Move detailed API descriptions and code snippets to Appendix.

\section{Meta-Learning Formulation and Adaptation Mechanisms}
\label{sec:meta_learning}

The adaptation task in the \thearena{} can be formalized as follows. Let $\mathcal{S} = \{s_1, s_2, \dots, s_N\}$ be the set of possible game stages, where each stage $s \in \mathcal{S}$ is defined by its board configuration, rule set, and initial state. Some stages are public ($s \in \mathcal{S}_{public}$), while others are private and unseen during development ($s \in \mathcal{S}_{private}$). An intelligent system $I$ is evaluated on its ability to adapt to a given stage $s$.

For each stage $s$, the intelligent system $I$ is granted a limited analysis time $T_{\text{analysis}} \approx 60$ seconds. During this time, $I$ can interact with the environment defined by $s$ solely through a predefined API (as described in Section~\ref{sub:api_interaction}). The goal of the system during this phase is to output a task-specific strategy function $f_s$:
\begin{equation}
I(s, \text{API}) \xrightarrow{T_{\text{analysis}}} f_s
\end{equation}
This function $f_s$ takes the current board state $b$, player ID $p$, and set of valid moves $M$ as input, and must return a chosen move. The cumulative time taken by $f_s$ across all its calls within a single game is limited to $T_{\text{game}} \approx 10$ seconds:
\begin{equation}
\sum_{\text{moves in game}} \text{Time}(f_s(b, p, M) \rightarrow m) \le T_{\text{game}}
\end{equation}

The performance of the intelligent system $I$ on stage $s$ is evaluated based on the effectiveness of the generated strategy $f_s$ in playing games on stage $s$. The overall evaluation score for system $I$ is a composite metric considering various factors across a set of evaluation stages $\mathcal{S}_{eval} \subseteq \mathcal{S}_{private}$. Key metrics (Section \ref{sub:metrics}) are Task Performance ($P$), Adaptation Speed ($A$), Efficiency ($E$), Generalization ($G$), and Robustness ($R$). This aligns with few-shot adaptation~\cite{brown2020language} under tight constraints.

\subsection{Adaptation Strategies under Time Constraints}
\label{sub:adaptation_strategies}

Effectively addressing the \thearena{} challenge requires intelligent systems to devise strategies for analyzing a new environment and generating a tailored strategy within the $T_{\text{analysis}}$ time limit. This involves an exploration-exploitation trade-off: how much time should be spent exploring the new environment's dynamics versus optimizing the strategy based on the acquired knowledge?

Successful adaptation mechanisms in this context often involve a combination of techniques as:

\begin{itemize}
  \item \textbf{Environmental Modeling through Interaction:} Since the intelligent system does not have explicit knowledge of the specific rule or structural variations of a novel stage beyond what is revealed through interaction via the API, it must learn an implicit or explicit model of that environment's dynamics. A common approach, demonstrated in \texttt{intelligent-system-template-adv.js}, is to perform rapid self-play simulations using the provided API functions (\texttt{getValidMoves}, \texttt{simulateMove}, \texttt{evaluateBoard}). By playing a large number of simulated games ($\approx 3000$ games in the template) within the $T_{\text{analysis}}$ window, the system can:\footnote{Notably, this level of simulation ($\approx 3000$ games) is significantly higher than what a human typically requires (e.g., 3-4 games) to grasp the rules and achieve intermediate play, highlighting a current difference in learning efficiency between such systems and human adaptability~\cite{lake2017building}.}
  \begin{itemize}
    \item Infer the rules of piece movement and capture, including any variations like `ignore occlusion' or `fewer pieces continue' (as shown by the rule detection logic in the template).
    \item Gather statistics on the value of different board positions or move sequences in the context of the current stage (e.g., calculating position win rates in the template).
    \item Understand the typical game flow and identify effective opening sequences for this specific environment.
  \end{itemize}
  \item \textbf{Learning Environmental Features:} Based on the data gathered through simulation or initial board inspection, the system can identify key features of the environment, such as board size (\texttt{stageConfig.boardSize}), presence of blocked cells (\texttt{hasBlockedCells} in the code snippet), or detected rule variations. These features inform the subsequent strategy generation.
  \item \textbf{Adaptive Strategy Synthesis:} Using the learned environmental model and identified features, the intelligent system synthesizes a strategy function $f_s$. This can involve:
  \begin{itemize}
    \item Adjusting parameters of a general Othello strategy template (e.g., modifying positional weights or search depth based on board size or blocked cells).
    \item Selecting from a portfolio of pre-existing strategy components or heuristics based on the detected environmental features.
    \item Constructing a new strategy using learned elements, such as a data-driven positional value matrix (\texttt{positionValueMatrix}) or an opening book (\texttt{openingBook}) derived from the self-play simulations in the current stage.
  \end{itemize}
  \item \textbf{Efficient Time Allocation:} The $T_{\text{analysis}}$ constraint necessitates a trade-off between the depth of environmental exploration (e.g., number of simulations) and the complexity of the strategy synthesis process. Successful systems are expected to adaptively manage the total $T_{\text{game}}$ budget, spending more time exploring highly novel environments and perhaps relying on very fast heuristics or shallow search for familiar moves to stay within the cumulative 10-second limit. 
\end{itemize}

This process requires a meta-cognitive ability to understand the task at hand (the specific stage), assess what information is needed for effective play, and efficiently acquire and utilize that information within the given time budget.

\subsection{Connection to General Intelligence}
\label{sub:agi_connection}

The challenges posed by the \thearena{} are deeply connected to the core principles of general intelligence. True AGI is not merely about excelling at a single task but demonstrating robust performance across a wide range of tasks and the capacity to adapt to novel ones.

\begin{itemize}
    \item \textbf{Rapid Task Acquisition:} The limited-time adaptation challenge directly evaluates a system's ability to quickly understand and become proficient in a new task (stage), a fundamental aspect of intelligent behavior.
    \item \textbf{Environmental Modeling:} Adapting to unseen rule and structural variations requires building or updating an internal model of the environment's dynamics. The simulation-based learning approach in \texttt{intelligent-system-template-adv.js} is one method for achieving this.
    \item \textbf{Transfer Learning and Generalization:} Effective intelligent systems should leverage knowledge gained from public stages or prior experience to accelerate adaptation on private stages~\cite{reed2022gato}. This requires abstracting principles that generalize across different Othello variants.
    \item \textbf{Meta-Cognition and Resource Management:} The time constraints compel systems to reason about their own cognitive processes – deciding how to allocate the $T_{\text{analysis}}$ budget, when to stop exploring, and which learning methods are most efficient for the given stage.
    \item \textbf{Flexible Strategy Synthesis:} Instead of relying on fixed algorithms, intelligent systems must be able to compose or modify strategies based on learned environmental properties, demonstrating strategic flexibility.
\end{itemize}

By providing a controlled yet challenging environment for evaluating these capabilities, the \thearena{} serves as a valuable microcosm for studying and advancing the development of general intelligent agents.

\section{Evaluation Methodology and Dataset Potential}
\label{sec:evaluation}

Evaluating adaptive intelligence requires metrics and methodologies that go beyond traditional performance measures in static environments. The \thearena{} employs a multi-faceted approach to assess the capabilities of submitted intelligent systems and offers the potential for a valuable dataset for future research.

\subsection{Evaluation Metrics}
\label{sub:metrics}

Performance of intelligent systems is evaluated based on several key dimensions, designed to capture different facets of adaptive intelligence:

\begin{itemize}[leftmargin=*, label=\textbullet] 
    \item \textbf{Task Performance ($P$):} This is the most straightforward metric, measuring the success of the generated strategy function ($f_s$) in playing games on the target stage $s$. It is typically quantified by win rate, average score difference against various opponents (including built-in strategies like Random, Greedy, Corners, Positional, and potentially other submitted intelligent systems), and the number of pieces controlled at the end of the game.
 \item \textbf{Adaptation Speed ($A$):} This measures how quickly an intelligent system is able to generate an effective strategy for a \textit{new} stage. In the context of the 60-second analysis limit, adaptation speed can be implicitly assessed by the performance of the generated strategy in the initial games played on a private stage. A system that rapidly identifies key environmental features and generates a reasonable strategy will likely perform better early on than one that is slow to understand the new environment. Future metrics could involve explicitly measuring performance within a fixed number of initial moves or games on the new stage.
 \item \textbf{Efficiency ($E$):} This dimension evaluates the computational cost of the intelligent system within the imposed time limits. It includes:
 \begin{itemize}[leftmargin=*, label=-]
   \item \textit{Analysis Time Utilization:} How effectively the system uses the allocated $T_{\text{analysis}}$ ($\approx 60$ seconds). This is measured by the actual time taken by the \texttt{analyzeStage} function. Systems that exceed this limit fail to produce a strategy.
   \item \textit{Game Time Utilization:} How effectively the generated strategy function $f_s$ uses its \textit{total game time budget of $\approx 10$ seconds} across all moves in a single game. Exceeding this cumulative limit results in a time forfeit. This constraint necessitates fast individual move computations, typically on the order of tens or a couple of hundred milliseconds per move.
 \end{itemize}
 \item \textbf{Generalization ($G$):} A critical measure of adaptive intelligence is the ability to perform well on unseen tasks. Generalization is assessed by comparing the performance of intelligent systems on the public training stages versus the private evaluation stages. A small performance drop on private stages indicates strong generalization capabilities.
 \item \textbf{Adaptation Robustness ($R$):} This is evaluated by testing the system's consistent performance across a \textit{variety} of different, challenging stages, including those with combinations of structural and rule variations. A system that maintains strong performance across diverse unseen environments demonstrates high adaptation robustness.
\end{itemize}

A final evaluation score for an intelligent system is computed as a weighted combination of these dimensions, allowing the benchmark to highlight systems that are not just strong players but are also fast, efficient, and robust adaptors to novel environments.

Central to the philosophy of the \thearena{}, and aligned with definitions of intelligence emphasizing skill-acquisition efficiency (e.g., \cite{chollet2019ARC}), is the concept that intelligence is demonstrated not merely by high performance (skill), but by the efficiency with which that skill is acquired and generalized to novel situations, controlling for prior knowledge and experience. From this perspective, a system that relies heavily on a large amount of pre-encoded prior knowledge tailored to potential variations might be considered less intelligent than one that can achieve comparable performance and generalization with less prior knowledge, by more effectively learning and adapting from limited exposure to the new environment.

This understanding motivates potential benchmark design choices, such as imposing limits on the size of the intelligent system (as a proxy for the amount of prior knowledge it contains), or restricting access to external information sources or powerful pre-trained models (like large language models) during the time-constrained analysis phase. Similar to approaches seen in benchmarks like the ARC-prize, such constraints would further emphasize the system's ability to generalize and adapt efficiently based on its intrinsic capabilities and the limited information available within the benchmark's environment, rather than relying on extensive external resources.

\subsection{Tournament Setup}
\label{sub:tournament_setup}

The evaluation of intelligent systems and their generated strategies is conducted using an automated tournament system (\texttt{tournament.js}). While referred to as a tournament, the evaluation on each stage follows a round-robin format. Submitted intelligent systems analyze each private stage sequentially within the $T_{\text{analysis}}$ limit, producing a tailored strategy function ($f_s$) for that specific stage. These generated strategies then compete against each other and against a set of baseline built-in strategies (e.g., Random, Greedy, Positional from \texttt{strategies.js}), as well as various advanced strategies (including those inspired by Logistello~\cite{buro2019logistello}), on the corresponding stage.

For each stage in the evaluation set:
\begin{enumerate}
 \item Each intelligent system's \texttt{analyzeStage} function is executed with the stage configuration and API, strictly limited to $T_{\text{analysis}}$.
 \item If a strategy function $f_s$ is successfully generated within the time limit, it is entered into the round-robin competition for stage $s$.
 \item In the round-robin, each generated strategy plays against every other strategy (including baselines and advanced strategies) as both Black and White.
 \item Strategies compete in games on stage $s$, with each strategy function's total time for the entire game limited to $T_{\text{game}} \approx 10$ seconds.
 \item Match results (win/loss/draw, final scores) are recorded by the logging system (\texttt{game-logger.js}).
 \item Performance metrics ($P, A, E$) are calculated for each generated strategy on stage $s$.
\end{enumerate}
After evaluating systems across all private stages, the Generalization ($G$) and Robustness ($R$) metrics are computed. The final leaderboard, prominently displayed on the web interface (Top right corner of Fig~\ref{fig:arena_screenshot} illustrates a sample leaderboard display), is then generated based on the aggregated weighted scores, with the Task Performance ($P$) metric serving as the primary visual representation of evaluation results for each strategy on a given stage.

\subsection{Dataset Structure and Potential}
\label{sub:dataset}

The \thearena{} platform is designed to generate a rich and unique dataset for research into adaptive AI. A key feature enabling this is the "Save Log" function, which allows users to export comprehensive data from individual games or full tournaments for offline analysis. This data serves as a valuable record of intelligent system behavior and environmental interactions under various conditions. For every evaluation run of an intelligent system on a stage, the system captures detailed information, including:

\begin{itemize}[leftmargin=*, label=\textbullet]
 \item \textbf{Stage Configuration}: Full details of the environment presented.
 \item \textbf{Intelligent System Code}: Source code capturing the adaptive logic.
 \item \textbf{Generated Strategy Code}: Source code of the task-level strategy function.
 \item \textbf{Analysis Process Logs \& Time}: Outputs and time usage during the $T_{\text{analysis}}$ phase, offering insight into the analysis and learning process.
 \item \textbf{Game Logs \& Results}: Detailed records of gameplay, including move sequences, board states, actions, time usage per move, and final outcomes (\texttt{game-logger.js}).
\end{itemize}

The detailed game data can be exported in two formats: a human-readable text log and a structured JSON format for automated analysis. See Appendix for examples and further details.

\section{Benchmark Insights from Preliminary Experiments}
\label{sec:results}

Full-scale experimental results involving a cohort of diverse intelligent systems across the complete set of private stages are currently being collected as part of an ongoing educational course. However, preliminary observations from pilot tests with prototype intelligent systems (e.g., variants of \texttt{intelligent-system-template-adv.js}) and received questions provide valuable insights into the types of results the \thearena{} yields and the patterns of adaptation observed.

\subsection{Illustrative Tournament Results}
\label{sub:illustrative_results}

Figure \ref{fig:arena_screenshot} shows results from a pilot tournament on "8x8 (Partial C-Squares-cw)", including built-in, advanced, and a prototype intelligent system's generated strategy. Strategies are ranked by Task Performance ($P$)---Win Rate.

\begin{itemize}[leftmargin=*, label=\textbullet]
  \item Simple strategies like Random and Greedy show relatively lower performance on the leaderboard.
    \item More sophisticated fixed strategies like smart-lv2 (from \texttt{smart-lv2.js}, having alpha-beta pruning with reasonable MCTS depth, and some opening books with robust heuristics) perform better.
    \item The strategy generated by the prototype intelligent system is competitive, outperforming some fixed strategies but not necessarily the best-performing ones. This highlights the challenge: generating a near-optimal strategy for a novel environment within a strict time limit is difficult, but achievable to a degree that surpasses non-adaptive or poorly generalizing fixed strategies.
\end{itemize}
This illustrative example demonstrates how the \thearena{} provides quantitative comparisons of strategy performance on specific, potentially unseen, environmental variations.

\subsection{Preliminary Observations on Adaptation Patterns}
\label{sub:adaptation_patterns}

Beyond win/loss records, the \thearena{} enables deeper analysis of adaptation. Preliminary observations from pilot tests suggest several patterns:

\begin{itemize}[leftmargin=*, label=\textbullet]
 \item \textbf{Time Allocation:} Systems varied in using the $T_{\text{analysis}}$ budget. A balanced approach between environmental modeling (simulation) and strategy synthesis likely correlates with better adaptation.
 \item \textbf{Game Time Limit Impact:} The strict $T_{\text{game}}$ limit ($\approx 10$s total) constrains computationally intensive strategies (like deep MCTS), favoring efficient, adaptive heuristics.
 \item \textbf{Simulation-Based Learning:} Self-play simulations within $T_{\text{analysis}}$ improved robustness to structural variations by learning stage-specific values.
 \item \textbf{Rule Adaptation Challenges:} Adapting to subtle rule variations appeared more challenging than structural changes, often requiring explicit detection logic or highly flexible simulation.
 \item \textbf{Generalization Gap:} Fixed strategies showed larger performance drops on private vs. public stages than adaptive systems, illustrating the specialization/generalization trade-off.
\end{itemize}
These preliminary findings demonstrate the benchmark's ability to reveal distinct adaptation patterns under time constraints.

\subsection{Potential for Analysis and Extension}
\label{sub:analysis_extension}

The detailed game logs and analysis outputs collected by the \thearena{} (as described in Section \ref{sub:dataset}) enable extensive post-hoc analysis of adaptation processes. Researchers can analyze correlations between analysis behaviors and strategy quality, system strategy adjustments, the relationship between strategy complexity and performance, and successes/failures in adapting to variations through log inspection. This transparency is a key strength supporting fine-grained analysis.

Building upon this framework, the \thearena{} is highly extensible and offers numerous avenues for future work and more complex challenges. These potential expansions aim to explore broader concepts of generalization and more sophisticated adaptive mechanisms. This could involve introducing challenges with incomplete information, such as limited observation or reasoning under uncertainty, or exploring multi-agent interaction scenarios, like N-Player or cooperative games. Further complexity could be added through dynamic environments where rules or configurations change during gameplay, or by requiring cross-domain adaptation to different game mechanics or environments. Additionally, the evaluation itself could evolve to include adaptive elements, such as variable time limits or automated curriculum generation based on system performance. These directions provide a roadmap for increasing benchmark complexity and pushing the boundaries of intelligent adaptation research.

\subsection{Conclusion}
\label{sec:conclusion}

In conclusion, this paper introduces the Othello AI Arena, a novel benchmark designed to address the critical gap in evaluating AI systems' capacity for limited-time adaptation to unseen environments. I frame this challenge as a meta-learning problem, where participants develop intelligent systems capable of analyzing a novel Othello stage configuration and rules within a strict time limit ($T_{\text{analysis}} \approx 60$s) to generate a tailored, high-performing strategy. The benchmark explicitly evaluates the meta-level intelligence of the system, separating it from the task-level performance of the generated strategy. The Othello AI Arena platform supports this higher-level competition, enabling the submission and evaluation of adaptive intelligent systems through tournaments with multi-dimensional metrics. While preliminary results offer initial insights into adaptation patterns, the comprehensive data generated also provides a valuable resource for analyzing the adaptation process itself. By focusing on rapid, intelligent adaptation under realistic constraints, the Othello AI Arena serves as a valuable tool for fostering and evaluating progress towards artificial general intelligence.

\bibliography{ARC_reference}
\bibliographystyle{plain}


\clearpage 

\appendix

\clearpage 

\setlength{\cftbeforesecskip}{1.0\baselineskip}
\setlength{\cftbeforesubsecskip}{1.0\baselineskip}
\setlength{\cftaftertoctitleskip}{1.0\baselineskip}

\section{System Architecture}
\label{app:system-architecture}

The Othello AI Arena's modular architecture is designed for clarity, maintainability, and extensibility, facilitating the development and evaluation of intelligent systems. Figure~\ref{fig:othello-arch} presents a comprehensive overview of the system's key components and their hierarchical interactions, with a specific focus on how data and control flow through the various layers.

The architecture is broadly divided into several logical layers: the \emph{Core Layer} handling fundamental game logic, the \emph{UI Layer} for rendering and interaction, the \emph{Strategy Layer} managing AI behaviors, and the \emph{Competition Layer} orchestrating tournaments. These layers are all orchestrated by the main \texttt{app.js} and rely on a \emph{Data Layer} for configurations and an \emph{Entry Point} for application loading. This layered approach allows participants to focus on developing the Intelligent System (primarily interacting with the Strategy Layer) while the platform handles underlying complexities.

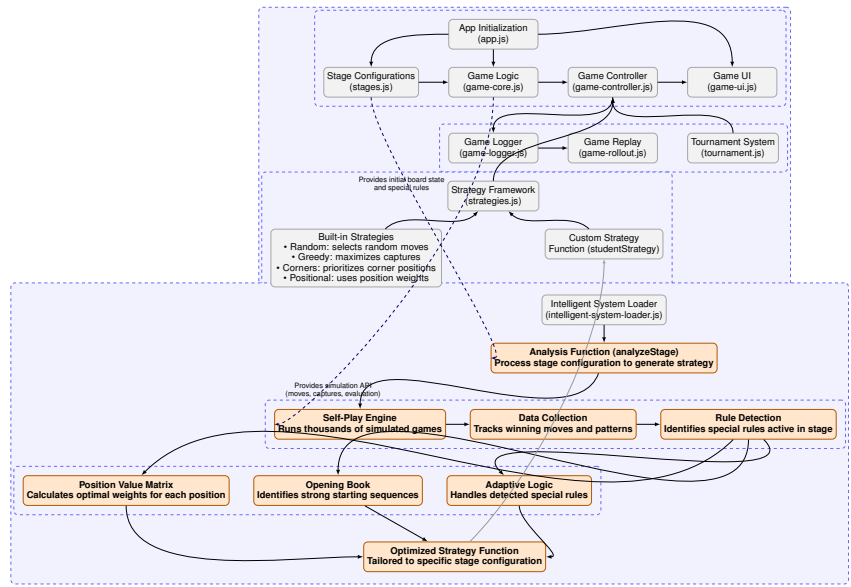
\begin{figure}[h]
   \centering
   \resizebox{0.8\textwidth}{!}{\begin{tikzpicture}[
    node distance=0.8cm,
    box/.style={
        rectangle, 
        draw=gray!70, 
        rounded corners, 
        minimum width=3cm, 
        minimum height=1cm, 
        text centered, 
        font=\sffamily\small,
        fill=gray!10,
        align=center
    },
    highlighted/.style={
        rectangle, 
        draw=orange!70!black, 
        rounded corners, 
        minimum width=3cm, 
        minimum height=1cm, 
        text centered, 
        font=\sffamily\small\bfseries,
        fill=orange!20,
        align=center,
        thick
    },
    group/.style={
        rectangle, 
        draw=blue!60, 
        dashed, 
        rounded corners, 
        inner sep=0.3cm,
        fill=blue!5,
        label={[font=\sffamily\small\bfseries]north:#1}
    },
    connectarrow/.style={->, >=latex, thick, draw=black},
    longconnect/.style={->, >=latex, thin, draw=black!40},
    dashedarrow/.style={->, >=latex, dashed, thin, draw=blue!40!black},
    label/.style={font=\sffamily\scriptsize, align=center}
]

\node[box] (app) {App Initialization\\(app.js)};
\node[box, below=0.6cm of app] (gamecore) {Game Logic\\(game-core.js)};
\node[box, right=1cm of gamecore] (gamecontrol) {Game Controller\\(game-controller.js)};
\node[box, right=1cm of gamecontrol] (gameui) {Game UI\\(game-ui.js)};
\node[box, left=1cm of gamecore] (stages) {Stage Configurations\\(stages.js)};

\node[box, below=1.2cm of gamecore] (logger) {Game Logger\\(game-logger.js)};
\node[box, right=1cm of logger] (replay) {Game Replay\\(game-rollout.js)};
\node[box, right=1cm of replay] (tournament) {Tournament System\\(tournament.js)};

\node[box, below=0.6cm of logger] (strategyfw) {Strategy Framework\\(strategies.js)};
\node[box, below left=0.6cm and 0.2cm of strategyfw, text width=5.5cm] (builtin) {Built-in Strategies\\
\textbullet~Random: selects random moves\\
\textbullet~Greedy: maximizes captures\\
\textbullet~Corners: prioritizes corner positions\\
\textbullet~Positional: uses position weights};
\node[box, below right=0.6cm and 0.2cm of strategyfw] (customstrat) {Custom Strategy\\Function (studentStrategy)};

\node[box, below=1.2cm of customstrat] (isloader) {Intelligent System Loader\\(intelligent-system-loader.js)};
\node[highlighted, below=0.6cm of isloader] (analyzefunc) {Analysis Function (analyzeStage)\\Process stage configuration to generate strategy};

\node[highlighted, below left=1.2cm and 1.5cm of analyzefunc] (selfplay) {Self-Play Engine\\Runs thousands of simulated games};
\node[highlighted, right=0.8cm of selfplay] (datacollect) {Data Collection\\Tracks winning moves and patterns};
\node[highlighted, right=0.8cm of datacollect] (ruledetect) {Rule Detection\\Identifies special rules active in stage};

\node[highlighted, below left=1.2cm and 1.5cm of selfplay] (valuematrix) {Position Value Matrix\\Calculates optimal weights for each position};
\node[highlighted, right=0.8cm of valuematrix] (openingbook) {Opening Book\\Identifies strong starting sequences};
\node[highlighted, right=0.8cm of openingbook] (adaptlogic) {Adaptive Logic\\Handles detected special rules};

\node[highlighted, below right=1.2cm and -2cm of openingbook] (outputstrategy) {Optimized Strategy Function\\Tailored to specific stage configuration};

\begin{pgfonlayer}{background}
    \node[group={Othello Arena Application}, fit=(app) (gamecore) (gamecontrol) (gameui) (stages) (logger) (replay) (tournament) (strategyfw) (builtin) (customstrat), inner sep=0.4cm] (arena) {};
    
    \node[group={Core Components}, fit=(app) (gamecore) (gamecontrol) (gameui) (stages), inner sep=0.3cm] (core) {};
    \node[group={Support Systems}, fit=(logger) (replay) (tournament), inner sep=0.3cm] (support) {};
    \node[group={Strategy System}, fit=(strategyfw) (builtin) (customstrat), inner sep=0.3cm] (strategysystem) {};
    
    \node[group={Student Assignment: Intelligent System}, fit=(isloader) (analyzefunc) (selfplay) (datacollect) (ruledetect) (valuematrix) (openingbook) (adaptlogic) (outputstrategy), inner sep=0.4cm] (is) {};
    
    \node[group={Simulation Phase}, fit=(selfplay) (datacollect) (ruledetect), inner sep=0.3cm] (simphase) {};
    \node[group={Strategy Generation Phase}, fit=(valuematrix) (openingbook) (adaptlogic), inner sep=0.3cm] (genphase) {};
\end{pgfonlayer}

\draw[connectarrow] (app) -- (gamecore);
\draw[connectarrow] (app) to[out=180, in=90, looseness=1] (stages);
\draw[connectarrow] (app) to[out=0, in=90, looseness=1] (gameui);
\draw[connectarrow] (gamecore) -- (gamecontrol);
\draw[connectarrow] (gamecontrol) -- (gameui);
\draw[connectarrow] (stages) -- (gamecore);

\draw[connectarrow] (gamecontrol) to[out=270, in=90, looseness=0.8] (logger);
\draw[connectarrow] (logger) -- (replay);
\draw[connectarrow] (tournament) to[out=90, in=270, looseness=0.8] (gamecontrol);

\draw[connectarrow] (strategyfw) to[out=90, in=270, looseness=0.8] (gamecontrol);
\draw[connectarrow] (builtin) to[out=45, in=225, looseness=0.8] (strategyfw);
\draw[connectarrow] (customstrat) to[out=135, in=315, looseness=0.8] (strategyfw);

\draw[connectarrow] (isloader) -- (analyzefunc);
\draw[connectarrow] (analyzefunc) to[out=250, in=90, looseness=0.8] (selfplay);
\draw[connectarrow] (selfplay) -- (datacollect);
\draw[connectarrow] (datacollect) -- (ruledetect);

\draw[connectarrow] (ruledetect) to[out=270, in=90, looseness=0.8] (openingbook);
\draw[connectarrow] (ruledetect) to[out=225, in=45, looseness=0.8] (valuematrix);
\draw[connectarrow] (ruledetect) to[out=315, in=135, looseness=0.8] (adaptlogic);

\draw[connectarrow] (valuematrix) to[out=270, in=180, looseness=0.8] (outputstrategy);
\draw[connectarrow] (openingbook) -- (outputstrategy);
\draw[connectarrow] (adaptlogic) to[out=270, in=0, looseness=0.8] (outputstrategy);

\draw[longconnect] (outputstrategy) to[out=45, in=270, looseness=0.7] (customstrat);

\draw[dashedarrow] (stages) .. controls +(0,-2) and +(-3,0) .. node[label, pos=0.3, below] {Provides initial board state\\and special rules} (analyzefunc);
\draw[dashedarrow] (gamecore) .. controls +(0,-5) and +(-2,0) .. node[label, pos=0.7, below] {Provides simulation API\\(moves, captures, evaluation)} (selfplay);

\end{tikzpicture}}
   \caption{System Architecture of the Othello AI Arena (Gray part) and its interaction with the potential intelligent system (Orange part).}
   \label{fig:architecture}
\end{figure}

\section{Core Modules and Implementation Details}
\label{app:implementation-details}

The Othello AI Arena is built upon a modular JavaScript architecture, separating concerns into distinct modules that interact through well-defined interfaces. This design facilitates maintainability, extensibility, and clarity in understanding the system's operation. Below, I detail the core modules that underpin the arena's functionality, illustrating their responsibilities and key code components.

\subsection{Game Core Engine}
\label{app:game-core}
The \texttt{game-core.js} module is the heart of the Othello Arena, implementing the fundamental game rules and variations. It manages the board state, validates moves, executes piece captures, and determines the next player based on the active stage's rules. This module is critical as its behavior (e.g., how \texttt{isValidMove} or \texttt{makeMove} function under different rule variations) is what intelligent systems must infer through observation via the API.

\begin{itemize}
    \item \textbf{Key Responsibilities}: Board initialization and management, move validation (\texttt{isValidMove}), move execution (\texttt{makeMove}), piece counting (\texttt{countDiscs}), and determining the next player (\texttt{determineNextPlayer}).
    \item \textbf{Illustrative Code Snippet (Move Execution Logic):} This snippet demonstrates how a move is executed, involving placing a piece and flipping captured pieces. The internal logic for flipping pieces must implicitly handle stage-specific rules like `ignore occlusion' without exposing these rules to the intelligent system API.
\end{itemize}

\begin{minted}[linenos, fontsize=\small, frame=lines, breaklines=true]{javascript}
// Excerpt from game-core.js: Core move execution logic
function makeMove(row, col, player) {
    if (!isWithinBoard(row, col) || board[row][col] !== GAME_CONSTANTS.EMPTY) return false;
    board[row][col] = player; // Place the player's piece

    const opponent = player === GAME_CONSTANTS.BLACK ? GAME_CONSTANTS.WHITE : GAME_CONSTANTS.BLACK;
    const directions = [[-1, -1], [-1, 0], [-1, 1], [0, -1], [0, 1], [1, -1], [1, 0], [1, 1]];

    // Internal rule application based on currentStage configuration (not visible to AI via API)
    const stageConfig = currentStage || stages[0]; // Internal to game-core
    const ignoreOcclusion = stageConfig.ignoreOcclusion || false; // Internal flag

    const capturedPieces = [];
    // Logic to find and flip pieces, considering ignoreOcclusion if true
    // The AI observes the result via simulateMove, inferring the rule.
    for (const [dr, dc] of directions) {
        let r = row + dr;
        let c = col + dc;
        const toFlip = [];
        let foundBlocked = false;
        let foundOpponent = false;

        while (isWithinBoard(r, c)) {
            if (board[r][c] === opponent) {
                toFlip.push([r, c]);
                foundOpponent = true;
            } else if (board[r][c] === GAME_CONSTANTS.BLOCKED) {
                foundBlocked = true;
                if (!ignoreOcclusion) { // If occlusion is not ignored, break
                    break;
                }
            } else if (board[r][c] === GAME_CONSTANTS.EMPTY) {
                break;
            } else if (board[r][c] === player) {
                if (foundOpponent && toFlip.length > 0 && (!foundBlocked || ignoreOcclusion)) {
                    for (const [fr, fc] of toFlip) {
                        board[fr][fc] = player;
                        capturedPieces.push([fr, fc]);
                    }
                }
                break;
            }
            r += dr;
            c += dc;
        }
    }
    // ... (rest of makeMove function, e.g., logging) ...
    return true; // Or false if move was invalid
}
\end{minted}

\textbf{Note:} The \texttt{ignoreOcclusion} flag is an internal detail of the \texttt{game-core.js} module, used by the platform to define the specific behavior of \texttt{simulateMove} and \texttt{getValidMoves} for a given stage. Intelligent systems must infer the presence of such rules through observing the API's behavior, not by direct access to \texttt{stageConfig} flags.

\subsection{Intelligent System Execution Environment}
\label{app:intelligent-system-env}
The \texttt{intelligent-system-loader.js} module is responsible for securely loading, sandboxing, and executing submitted intelligent system code. It enforces the strict $T_\text{analysis}$ time limit for strategy generation and provides the limited API through which intelligent systems interact with the environment.

\begin{itemize}
    \item \textbf{Key Responsibilities}: Code compilation (\texttt{compileSystem}), secure execution in a Web Worker, time limit enforcement, and exposing the environment API (\texttt{getValidMoves}, \texttt{simulateMove}, \texttt{evaluateBoard}) to the intelligent system.
    \item \textbf{Illustrative Code Snippet (Worker-based Execution):} This shows the core mechanism for running student-submitted \texttt{analyzeStage} functions in a sandboxed environment, managing time, and passing the API.
\end{itemize}

\begin{minted}[linenos, fontsize=\small, frame=lines, breaklines=true]{javascript}
// Excerpt from intelligent-system-loader.js: Core analysis execution within a Worker
async analyzeStageWithSystem(systemName, stageConfig, systemCode) {
    return new Promise((resolve, reject) => {
        const worker = new Worker(URL.createObjectURL(new Blob([`
            self.onmessage = function(e) {
                const { code, stageConfig, initialBoard, validMoves, api } = e.data;
                const startTime = performance.now();
                const analyzeStage = new Function('stageConfig', 'initialBoard', 'validMoves', 'api',
                    code + '\nreturn analyzeStage.apply(null, arguments);'
                );
                try {
                    const strategyFunction = analyzeStage(stageConfig, initialBoard, validMoves, api);
                    self.postMessage({ status: 'success', strategyFunction: strategyFunction.toString() });
                } catch (error) {
                    self.postMessage({ status: 'error', message: error.message, stack: error.stack });
                }
            };
        `], { type: 'application/javascript' })));

        // Time limit enforced externally (in loader, not worker code shown here)
        const timeoutId = setTimeout(() => { worker.terminate(); reject(new Error("Analysis timed out")); }, this.timeLimit);

        worker.onmessage = (e) => {
            clearTimeout(timeoutId);
            if (e.data.status === 'success') { /* ... process strategy function ... */ }
            else { /* ... handle error ... */ }
        };
        worker.onerror = (error) => { clearTimeout(timeoutId); reject(error); };

        worker.postMessage({ code: systemCode, stageConfig, initialBoard, validMoves, api: this.api });
    });
}
\end{minted}

\subsection{Strategy Management}
\label{app:strategy-management}
The \texttt{strategies.js} module manages both built-in AI strategies and custom strategies uploaded by participants. It handles the storage, retrieval, and runtime compilation of strategy code.

\begin{itemize}
    \item \textbf{Key Responsibilities}: Defining standard Othello strategies (e.g., Random, Greedy, Positional), saving and loading custom strategy code (e.g., from \texttt{localStorage}), and providing \texttt{getCompiledStrategy} to retrieve an executable function for a given strategy ID. This module serves the \texttt{GameController} by providing the actual move-making functions.
    \item \textbf{Illustrative Code Snippet (Strategy Retrieval and Compilation):} This demonstrates how the arena retrieves an executable strategy, prioritizing cached compiled functions for efficiency.
\end{itemize}

\begin{minted}[linenos, fontsize=\small, frame=lines, breaklines=true]{javascript}
// Excerpt from strategies.js: Retrieving and compiling strategies
function getCompiledStrategy(controllerId, player) {
    // Handle custom strategies (student-submitted)
    if (controllerId.startsWith('custom_')) {
        const strategyName = controllerId.replace('custom_', '');
        // Prioritize cached compiled function (from intelligent-system-loader)
        if (window.compiledIntelligentSystems[strategyName]) {
            return window.compiledIntelligentSystems[strategyName]; // Function stored by intelligent-system-loader
        }
        // Fallback: compile from stored code string if not cached
        if (savedStrategies[strategyName]) {
            const code = savedStrategies[strategyName];
            try {
                const compiledFunc = new Function('boardArg', 'playerArg', 'validMovesArg', 'makeMoveFunc',
                    `${code}\nreturn studentStrategy(boardArg, playerArg, validMovesArg, makeMoveFunc);`);
                compiledStudentAIs[strategyName] = compiledFunc; // Cache it
                return compiledFunc;
            } catch (e) { console.error(`Compile error for ${strategyName}:`, e); return null; }
        }
    }
    // Handle built-in strategies
    else if (builtInStrategies[controllerId]) {
        return builtInStrategies[controllerId]; // Directly use built-in functions
    }
    return null;
}
\end{minted}

\subsection{Game Flow and UI Management}
\label{app:game-ui-flow}
The \texttt{game-controller.js} and \texttt{game-ui.js} modules work in tandem to orchestrate the game flow and manage the web-based user interface.

\begin{itemize}
    \item \textbf{GameController (\texttt{game-controller.js})}: Manages the overall game loop, including starting/resetting games, calling AI strategies for moves, and determining game-over conditions. It also tracks player time usage.
    \item \textbf{OthelloUI (\texttt{game-ui.js})}: Handles all aspects of the graphical user interface, including rendering the Othello board, updating scores and timers, displaying game status messages, and processing human player inputs.
\end{itemize}

\subsection{Logging and Replay System}
\label{app:logging-replay}
The \texttt{game-logger.js} and \texttt{game-rollout.js} modules are dedicated to recording game history and enabling post-game analysis and visualization.

\begin{itemize}
    \item \textbf{GameLogger (\texttt{game-logger.js})}: Provides a comprehensive system for recording every move, board state, player turn, and captured piece count during a game. It also stores game results for the leaderboard and supports saving data to \texttt{localStorage}.
    \item \textbf{GameRollout (\texttt{game-rollout.js})}: Utilizes the data from \texttt{GameLogger} to allow users to replay any past game, navigate through moves, and adjust playback speed. It dynamically reconstructs the board state at any point in the game.
\end{itemize}

\subsection{Tournament System}
\label{app:tournament-system}
The \texttt{tournament.js} module orchestrates automated tournaments between submitted intelligent systems and built-in strategies.

\begin{itemize}
    \item \textbf{Key Responsibilities}: Managing tournament rounds, recording match outcomes, calculating and displaying the leaderboard, and saving/loading tournament data. It provides the competitive environment for evaluating adaptive performance across private stages.
\end{itemize}

\section{Othello Arena API Reference}

The Othello AI Arena provides the following core API functions for developing intelligent systems, allowing them to probe and understand the environment:

\begin{itemize}
    \item \texttt{getValidMoves(board, player)}: Returns a list of valid moves for a player on a given board state. This function dynamically applies the stage's underlying rules, meaning the intelligent system must observe its behavior to infer rule variations.
    \item \texttt{simulateMove(board, player, row, col)}: Simulates the outcome of making a specific move, returning the resulting board state and number of pieces captured. The behavior of this function (e.g., how pieces are captured) directly reflects the stage's rules, allowing the intelligent system to deduce capture mechanics without explicit rule disclosure.
    \item \texttt{evaluateBoard(board, player)}: Provides an evaluation of a given board state using pre-defined metrics (e.g., piece count, mobility, corner control). The intelligent system may need to learn or adjust the interpretation of these metrics based on observed game dynamics.
\end{itemize}

An intelligent system is implemented through the \texttt{analyzeStage} function, which must analyze the stage within a strict time limit (approximately 60 seconds) and return a tailored strategy function. The returned strategy function is then called during each turn of the game to determine a move.


\begin{minted}[linenos, fontsize=\small, frame=lines, breaklines=true]{javascript}
// Basic Intelligent System Template
function analyzeStage(stageConfig, initialBoard, validMoves, api) {
    // Perform stage analysis within the ~60-second limit
    console.log("Analyzing stage: " + stageConfig.name);

    // Identify static environmental features (e.g., board dimensions, presence of blocked cells)
    const boardSize = initialBoard.length;
    const hasBlockedCells = initialBoard.some(row => 
        row.some(cell => cell === GAME_CONSTANTS.BLOCKED)); // GAME_CONSTANTS.BLOCKED is a known constant

    // The core challenge: Adapt strategy parameters based on inferred rules and board properties.
    // This 'optimizeParameters' function (conceptual) would perform simulations and deductions.
    const params = optimizeParameters(initialBoard, api); 

    // Return the strategy function.
    // This function will be called during each game turn (cumulative ~10-second limit per game).
    return function(board, player, validMoves) {
        if (validMoves.length === 0) return null;
        
        // This function decides the best move for the current turn,
        // utilizing the parameters and understanding gained during analysis.
        return selectBestMove(board, player, validMoves, params);
    };
}
\end{minted}

Key patterns for effectively utilizing the API include:
\begin{enumerate}
    \item Thoroughly understanding board characteristics and inferring rule variations during the initial analysis phase through active probing via the API.
    \item Efficiently using \texttt{simulateMove} for game tree exploration and evaluation function optimization, which helps in implicitly learning the game's dynamic rules.
    \item Performing complex computations (like extensive simulations or rule induction) during the analysis phase and leveraging the results within the strategy function.
    \item Employing memoization and optimization techniques to minimize execution time during gameplay.
\end{enumerate}

\subsection{Example Intelligent System: Advanced Template}
This example demonstrates a more sophisticated intelligent system that performs self-play simulations to learn about the stage, detect rules, and generate an adaptive strategy. It showcases how the provided API functions can be leveraged within the analysis time limit ($T_{analysis}$) to synthesize a high-performing strategy for a novel environment.

\begin{minted}[linenos, fontsize=\small, frame=lines, breaklines=true]{javascript}
/**
 * Function that intelligent systems must implement to analyze a stage and generate a strategy
 *
 * @param {Object} stageConfig - Stage configuration object (contains name, boardSize, initial piece positions, but NO explicit rule flags).
 * @param {Array<Array<number>>} initialBoard - 2D array representing the initial board state.
 * @param {Array<Object>} validMoves - Valid moves for the first player (Black) on the initial board.
 * @param {Object} api - Environment API with methods:
 * - getValidMoves(board, player): Returns valid moves for a given board state.
 * - simulateMove(board, player, row, col): Returns the resulting board and captured count for a move.
 * - evaluateBoard(board, player): Provides evaluation metrics for a board.
 * (These API methods internally implement the stage's TRUE rules,
 * which the intelligent system must infer through observation).
 *
 * @returns {Function} Strategy function that will be called during gameplay.
 */
function analyzeStage(stageConfig, initialBoard, validMoves, api) {
    console.log("=== Starting Environmental Analysis and Strategy Adaptation ===");
    console.log("Stage:", stageConfig.name, "Board Size:", stageConfig.boardSize);

    // Setup simulation parameters for environmental probing and data collection
    const NUM_SIM_GAMES = 3000;
    const MAX_MOVES_PER_GAME = 60;
    const ANALYSIS_TIME_LIMIT_MS = 55000; // Leave a few seconds buffer for final strategy compilation

    // Track the start time to adhere to the T_analysis limit
    const startTime = Date.now();

    // Data structures for storing observations and learned patterns
    const gameSimulationLogs = []; // Stores detailed logs of simulated games
    const positionWinRates = {}; // Tracks win rates for specific positions
    const boardSize = stageConfig.boardSize;

    // Initialize position win rates tracking
    for (let r = 0; r < boardSize; r++) {
        for (let c = 0; c < boardSize; c++) {
            positionWinRates[`${r},${c}`] = { plays: 0, wins: 0, blackPlays: 0, blackWins: 0, whitePlays: 0, whiteWins: 0 };
        }
    }

    // Define simple, generic strategies for self-play simulations.
    // These strategies do not have prior knowledge of rule variations,
    // they just interact with the API's behavior.
    const simulationStrategies = {
        random: function (board, player, moves) {
            if (moves.length === 0) return null;
            return moves[Math.floor(Math.random() * moves.length)];
        },
        greedy: function (board, player, moves) {
            if (moves.length === 0) return null;
            let bestMove = moves[0];
            let maxCaptures = -1;

            for (const move of moves) {
                const result = api.simulateMove(board, player, move.row, move.col);
                if (result.valid && result.capturedCount > maxCaptures) {
                    maxCaptures = result.capturedCount;
                    bestMove = move;
                }
            }
            return bestMove;
        },
        // A simple positional heuristic, which might be adapted later
        basicPositional: function(board, player, moves) {
            if (moves.length === 0) return null;
            let bestMove = null;
            let bestScore = -Infinity;

            // Simple static positional values (can be adapted)
            const staticWeights = {
                '8': [
                    [120, -20, 20, 5, 5, 20, -20, 120],
                    [-20, -40, -5, -5, -5, -5, -40, -20],
                    [20, -5, 15, 3, 3, 15, -5, 20],
                    [5, -5, 3, 3, 3, 3, -5, 5],
                    [5, -5, 3, 3, 3, 3, -5, 5],
                    [20, -5, 15, 3, 3, 15, -5, 20],
                    [-20, -40, -5, -5, -5, -5, -40, -20],
                    [120, -20, 20, 5, 5, 20, -20, 120]
                ],
                '6': [
                    [50, -10, 10, 10, -10, 50],
                    [-10, -20, -5, -5, -20, -10],
                    [10, -5, 5, 5, -5, 10],
                    [10, -5, 5, 5, -5, 10],
                    [-10, -20, -5, -5, -20, -10],
                    [50, -10, 10, 10, -10, 50]
                ]
            };
            const weights = staticWeights[board.length] || staticWeights['8']; // Use 8x8 as default

            for (const move of moves) {
                const score = weights[move.row][move.col];
                if (score > bestScore) {
                    bestScore = score;
                    bestMove = move;
                }
            }
            return bestMove;
        }
    };

    // Run self-play simulation games to gather data and observe behaviors
    console.log(`Starting ${NUM_SIM_GAMES} simulation games (analysis time limit: ${ANALYSIS_TIME_LIMIT_MS}ms)...`);

    let gamesCompleted = 0;
    let totalMovesMade = 0;
    let analysisTimeoutReached = false;

    for (let gameNum = 0; gameNum < NUM_SIM_GAMES; gameNum++) {
        if (Date.now() - startTime > ANALYSIS_TIME_LIMIT_MS) {
            analysisTimeoutReached = true;
            console.warn(`Analysis time limit reached (${ANALYSIS_TIME_LIMIT_MS}ms). Stopping simulations.`);
            break;
        }

        const gameLog = {
            moves: [],
            finalBoard: null,
            winner: null, // 1: Black, 2: White, 0: Tie
            scores: { black: 0, white: 0 },
            turnTransitions: [] // To observe if turns consistently alternate or not
        };

        let currentBoard = initialBoard.map(row => [...row]); // Deep copy
        let currentPlayer = 1; // Black starts
        let movesInCurrentGame = 0;
        let consecutivePasses = 0;
        let gameOver = false;

        // Randomly select two simulation strategies for this game
        const simStrategyPlayer1 = Object.values(simulationStrategies)[Math.floor(Math.random() * Object.keys(simulationStrategies).length)];
        const simStrategyPlayer2 = Object.values(simulationStrategies)[Math.floor(Math.random() * Object.keys(simulationStrategies).length)];

        while (!gameOver && movesInCurrentGame < MAX_MOVES_PER_GAME) {
            if (Date.now() - startTime > ANALYSIS_TIME_LIMIT_MS) {
                analysisTimeoutReached = true;
                break;
            }

            const moves = api.getValidMoves(currentBoard, currentPlayer);

            if (moves.length === 0) {
                consecutivePasses++;
                if (consecutivePasses >= 2) {
                    gameOver = true; // Game over if both players pass
                    break;
                }
                gameLog.turnTransitions.push({ from: currentPlayer, to: (currentPlayer === 1 ? 2 : 1), pass: true });
                currentPlayer = (currentPlayer === 1 ? 2 : 1); // Switch player if current player has no moves
                continue;
            }

            consecutivePasses = 0; // Reset pass counter

            const chosenSimStrategy = (currentPlayer === 1) ? simStrategyPlayer1 : simStrategyPlayer2;
            const chosenMove = chosenSimStrategy(currentBoard, currentPlayer, moves);

            if (!chosenMove) { // Should ideally not happen if validMoves.length > 0
                gameOver = true;
                break;
            }

            const simResult = api.simulateMove(currentBoard, currentPlayer, chosenMove.row, chosenMove.col);

            if (simResult.valid) {
                const prevPlayer = currentPlayer;
                currentBoard = simResult.resultingBoard; // Update board
                gameLog.moves.push({ player: prevPlayer, position: chosenMove, capturedCount: simResult.capturedCount });

                // Update position win rates data
                const posKey = `${chosenMove.row},${chosenMove.col}`;
                if (positionWinRates[posKey]) {
                    positionWinRates[posKey].plays++;
                    if (prevPlayer === 1) {
                        positionWinRates[posKey].blackPlays++;
                    } else {
                        positionWinRates[posKey].whitePlays++;
                    }
                }

                // Determine next player based on observed behavior (this is where inference matters)
                const scores = countPieces(currentBoard); // Utility function below
                let nextPlayerCandidate = (prevPlayer === 1) ? 2 : 1; // Standard alternate

                // Attempt to infer 'fewer pieces continue' type rule by observing actual turn transitions
                // We're looking for patterns where the current player plays again
                const potentialValidMovesForNextPlayer = api.getValidMoves(currentBoard, nextPlayerCandidate);
                if (potentialValidMovesForNextPlayer.length === 0) { // If next player cannot move, current player might continue or game ends
                     // This simple check doesn't infer complex rules, but models standard game flow.
                     // A more advanced system would track actual 'next player' behavior based on api.getValidMoves and piece counts.
                     // For instance, if the game *does* switch players, but then switches back quickly when piece counts are skewed.
                }

                // If the game rules cause the same player to move again, the API's getValidMoves and simulateMove
                // will reflect that. The AI observes.
                // For this generic template, we simply switch player. If the actual game logic allows consecutive turns,
                // the `api.getValidMoves` for the *next* player would return an empty array if they were forced to pass,
                // and the *current* player would get another chance.
                // The key is that the AI *doesn't know* why it's happening, only *that* it's happening.
                currentPlayer = nextPlayerCandidate; // Proceed with standard turn alternation for simulation logic

                gameLog.turnTransitions.push({ from: prevPlayer, to: currentPlayer, pass: false, scoresAfterMove: scores });
                movesInCurrentGame++;
                totalMovesMade++;
            } else {
                console.warn("Invalid move in simulation (should not happen for valid moves)");
                gameOver = true;
                break;
            }
        }

        // Game finished - record final state and winner
        const finalScores = countPieces(currentBoard);
        gameLog.finalBoard = currentBoard;
        gameLog.scores = finalScores;

        if (finalScores.black > finalScores.white) { gameLog.winner = 1; }
        else if (finalScores.white > finalScores.black) { gameLog.winner = 2; }
        else { gameLog.winner = 0; } // Tie

        // Update position win rates based on game winner
        for (const move of gameLog.moves) {
            const posKey = `${move.position.row},${move.position.col}`;
            const posStats = positionWinRates[posKey];
            if (posStats && gameLog.winner === move.player) {
                posStats.wins++;
                if (move.player === 1) { posStats.blackWins++; } else { posStats.whiteWins++; }
            }
        }

        gameSimulationLogs.push(gameLog);
        gamesCompleted++;
    }

    console.log(`Completed ${gamesCompleted} games with ${totalMovesMade} total moves in ${Date.now() - startTime}ms`);

    // --- Phase 2: Environmental Behavior Analysis and Rule Inference ---
    // Instead of looking for specific rule names, we analyze observed behaviors for anomalies.
    const observedBehaviors = analyzeEnvironmentBehaviors(gameSimulationLogs, initialBoard, api);
    console.log("Observed Environmental Behaviors:", observedBehaviors);

    // --- Phase 3: Adaptive Strategy Synthesis ---
    // Generate a position value matrix based on observed win rates
    const adaptedPositionValueMatrix = generatePositionValueMatrix(positionWinRates, boardSize, observedBehaviors);

    // Create an opening book from successful simulated games
    const derivedOpeningBook = buildOpeningBook(gameSimulationLogs);

    // Return the final, adapted strategy function
    return createStrategyFunction(adaptedPositionValueMatrix, derivedOpeningBook, observedBehaviors, boardSize, api);

    // ======================== Utility Functions (Internal to analyzeStage) ========================

    // Counts pieces on a given board state
    function countPieces(board) {
        let black = 0, white = 0, empty = 0, blocked = 0;
        for (let r = 0; r < board.length; r++) {
            for (let c = 0; c < board[r].length; c++) {
                if (board[r][c] === 1) black++;
                else if (board[r][c] === 2) white++;
                else if (board[r][c] === 0) empty++;
                else if (board[r][c] === 3) blocked++;
            }
        }
        return { black, white, empty, blocked };
    }

    /**
     * Analyzes simulation logs to infer environmental behaviors (rules)
     * without knowing specific rule names beforehand.
     */
    function analyzeEnvironmentBehaviors(simLogs, initialBoard, api) {
        const behaviors = {
            // High-level observations of turn dynamics
            consecutiveTurnObserved: false,
            consecutiveTurnFrequency: 0,
            consecutiveTurnPlayerBias: { black: 0, white: 0 }, // If one player gets more consecutive turns

            // High-level observations of capture mechanics
            captureThroughBlockedCellsObserved: false, // Did a capture happen over a blocked cell?
            unusualCapturePatternObserved: false, // More general, did captures seem off?

            // Observations about winning conditions
            winConditionReversedObserved: false, // Did lowest score win unexpectedly?
            // ... add more abstract behavioral observations
        };

        let totalTurnTransitions = 0;
        let consecutiveTurnsCount = 0;
        let blackConsecutiveTurns = 0;
        let whiteConsecutiveTurns = 0;

        for (const game of simLogs) {
            if (game.moves.length < 2) continue;

            // Analyze turn dynamics
            for (let i = 1; i < game.moves.length; i++) {
                const prevPlayer = game.moves[i - 1].player;
                const currPlayer = game.moves[i].player;
                totalTurnTransitions++;
                if (prevPlayer === currPlayer) {
                    consecutiveTurnsCount++;
                    if (prevPlayer === 1) blackConsecutiveTurns++;
                    else whiteConsecutiveTurns++;
                }
            }

            // Analyze capture mechanics (basic heuristic)
            // Look for any capture that crosses a blocked cell in the final board state
            // This is still a simple heuristic; robust detection would need more targeted probes.
            for (const move of game.moves) {
                const startR = move.position.row;
                const startC = move.position.col;
                const player = move.player;
                const opponent = (player === 1) ? 2 : 1;

                // Re-simulate this move on the board *before* the move to check intermediate states
                // This is computationally intensive. A simpler approach is to look at final board and piece positions.
                // However, a real intelligent system would use API to actively probe
                // Create a board state before this move occurred if possible from logs, or approximate.

                // A more direct probing strategy:
                // Construct a small, controlled scenario to test a specific behavior pattern.
                // This is hard to do generically without knowing board structure.
                // e.g., if (initialBoard has blocked cells) { try to place a piece, then simulate. }
                // For instance: pick a random empty cell, place current player piece.
                // Does it flip pieces beyond a blocked cell?
                // This would be done with direct calls to api.simulateMove on crafted micro-boards.

                // Example: Try to detect 'capture through blocked cells' by crafting a minimal board and probing
                // This would need to happen at the very beginning of analyzeStage, before main loop.
                // This template will stick to observations from general self-play for simplicity.
                // A very simple check (less robust): Check if any flipped piece in the log has a blocked cell between it and the origin move.
                // This is difficult without reconstructing the exact path and board state for each capture.
            }

            // Analyze winning condition
            // If the game winner is the player with FEWER pieces, this suggests a reversed winning condition
            const finalScores = game.scores;
            if (finalScores.black > finalScores.white && game.winner === 2) { // Black has more pieces, but White wins
                behaviors.winConditionReversedObserved = true;
            } else if (finalScores.white > finalScores.black && game.winner === 1) { // White has more pieces, but Black wins
                behaviors.winConditionReversedObserved = true;
            }
        }

        if (totalTurnTransitions > 0) {
            behaviors.consecutiveTurnFrequency = consecutiveTurnsCount / totalTurnTransitions;
            if (behaviors.consecutiveTurnFrequency > 0.05) { // Threshold for significance
                behaviors.consecutiveTurnObserved = true;
                if (blackConsecutiveTurns > whiteConsecutiveTurns * 1.5) { // Significant bias towards black
                    behaviors.consecutiveTurnPlayerBias.black = blackConsecutiveTurns / consecutiveTurnsCount;
                } else if (whiteConsecutiveTurns > blackConsecutiveTurns * 1.5) { // Significant bias towards white
                    behaviors.consecutiveTurnPlayerBias.white = whiteConsecutiveTurns / consecutiveTurnsCount;
                }
            }
        }

        // Example: Probe for ignoreOcclusion at analysis start with a crafted board
        // This is a more direct test rather than observing during self-play.
        // It's still a heuristic, not guaranteed to find all types of hidden rules.
        const probeBoard = [
            [0,0,0,0,0,0,0,0],
            [0,0,0,0,0,0,0,0],
            [0,0,2,3,1,0,0,0], // Opponent, Blocked, Player (try to capture over blocked)
            [0,0,0,0,0,0,0,0],
            [0,0,0,0,0,0,0,0],
            [0,0,0,0,0,0,0,0],
            [0,0,0,0,0,0,0,0],
            [0,0,0,0,0,0,0,0]
        ];
        // Scale probe board to current board size if needed, or create a generic smaller one.
        // Assume probe board is compatible with boardSize or it's a fixed small test.

        // Simulate placing a piece that should capture over the blocked cell (if rules allow)
        // If current player is 1 (Black), they want to capture White (2) over Blocked (3).
        const simulatedProbeResult = api.simulateMove(probeBoard, 1, 2, 0); // Try to capture from (2,0) over (2,1),(2,2) to (2,4)
        if (simulatedProbeResult.valid && simulatedProbeResult.capturedCount > 0) {
             // If a capture occurred, and there's a blocked cell in the line between the start and end piece,
             // then it's likely 'ignoreOcclusion' or similar.
             // This needs more robust check: is there a blocked cell between the origin and the captured pieces?
             // A very simple proxy: if the move *would* be invalid in standard Othello due to a block, but here it's valid.
             // This requires knowing standard Othello rules internally.
             // For simplicity, we just mark it as "unusual capture pattern" if the probe yields unexpected captures.
             behaviors.captureThroughBlockedCellsObserved = true; // Still a simplification for the template
        }


        return behaviors;
    }


    // Generates a position value matrix, potentially adapted by observed behaviors
    function generatePositionValueMatrix(posRates, boardSize, behaviors) {
        const matrix = Array(boardSize).fill().map(() => Array(boardSize).fill(0));

        // Start with a generic positional weight matrix (can be a standard Othello heuristic)
        const baseWeights_8x8 = [
            [120, -20, 20, 5, 5, 20, -20, 120],
            [-20, -40, -5, -5, -5, -5, -40, -20],
            [20, -5, 15, 3, 3, 15, -5, 20],
            [5, -5, 3, 3, 3, 3, -5, 5],
            [5, -5, 3, 3, 3, 3, -5, 5],
            [20, -5, 15, 3, 3, 15, -5, 20],
            [-20, -40, -5, -5, -5, -5, -40, -20],
            [120, -20, 20, 5, 5, 20, -20, 120]
        ];
        const baseWeights_6x6 = [
            [50, -10, 10, 10, -10, 50],
            [-10, -20, -5, -5, -20, -10],
            [10, -5, 5, 5, -5, 10],
            [10, -5, 5, 5, -5, 10],
            [-10, -20, -5, -5, -20, -10],
            [50, -10, 10, 10, -10, 50]
        ];
        const baseWeights = (boardSize === 6) ? baseWeights_6x6 : baseWeights_8x8;

        for (let r = 0; r < boardSize; r++) {
            for (let c = 0; c < boardSize; c++) {
                matrix[r][c] = baseWeights[r] ? baseWeights[r][c] : 0; // Handle different board sizes if needed
            }
        }

        // Adjust based on observed win rates from simulations
        for (let r = 0; r < boardSize; r++) {
            for (let c = 0; c < boardSize; c++) {
                const posKey = `${r},${c}`;
                const posStats = posRates[posKey];

                if (posStats && posStats.plays >= 3) { // Require minimum plays for confidence
                    const confidence = Math.min(1, posStats.plays / 10);
                    const winRateAdjustment = (posStats.wins / posStats.plays - 0.5) * 100 * confidence;
                    matrix[r][c] = Math.round(matrix[r][c] * 0.7 + winRateAdjustment * 0.3); // Blend with base
                }
            }
        }

        // Adapt matrix based on observed behaviors
        if (behaviors.winConditionReversedObserved) {
            console.log("Adapting position matrix for reversed winning condition.");
            // Invert the values to prioritize positions that lead to fewer pieces
            for (let r = 0; r < boardSize; r++) {
                for (let c = 0; c < boardSize; c++) {
                    matrix[r][c] = -matrix[r][c];
                }
            }
        }
        // Add more adaptations based on other behaviors...

        return matrix;
    }

    // Builds an opening book based on successful simulated games
    function buildOpeningBook(simLogs) {
        const openings = {};

        simLogs.forEach(game => {
            if (game.moves.length < 4) return; // Opening sequence too short

            const openingMoves = game.moves.slice(0, Math.min(6, game.moves.length));
            const openingKey = openingMoves.map(m => `${m.player}:${m.position.row},${m.position.col}`).join('|');

            if (!openings[openingKey]) {
                openings[openingKey] = { sequence: openingMoves, wins: 0, losses: 0, games: 0 };
            }
            openings[openingKey].games++;
            if (game.winner === 1) openings[openingKey].wins++;
            else if (game.winner === 2) openings[openingKey].losses++;
        });

        // Select the best openings (e.g., top 3 by win rate, with minimum plays)
        const goodOpenings = Object.values(openings)
            .filter(o => o.games >= 2 && (o.wins / o.games) > 0.6)
            .sort((a, b) => (b.wins / b.games) - (a.wins / a.games))
            .slice(0, 3);

        return goodOpenings.map(o => ({
            sequence: o.sequence.map(m => ({ player: m.player, position: { row: m.position.row, col: m.position.col } })),
            winRate: o.games > 0 ? o.wins / o.games : 0
        }));
    }

    /**
     * Creates the final strategy function that the Othello Arena will execute during gameplay.
     * This function should be fast and rely on pre-computed data from the analysis phase.
     */
    function createStrategyFunction(positionValues, openingBook, behaviors, boardSize, api) {
        console.log("\n=== Final Strategy Function Created ===");
        console.log(`Leveraging ${openingBook.length} opening sequences and adapted position values.`);
        console.log("Adapting based on observed behaviors:", behaviors);

        return function (board, player, validMoves) {
            // Handle no valid moves (should be handled by GameController for passes)
            if (validMoves.length === 0) return null;

            // --- Phase 1: Opening Book (if applicable) ---
            const pieceCount = countPieces(board).black + countPieces(board).white;
            if (pieceCount < 12 && openingBook.length > 0) { // Still early game
                // Attempt to follow an opening sequence from the derived book
                for (const opening of openingBook) {
                    // This matching logic is simplified; a real system would hash board states
                    // or track sequence of moves to find exact matches.
                    // For now, it just checks if the *next move in an opening* is valid.
                    for (const moveData of opening.sequence) {
                        if (moveData.player === player) {
                            const openingMove = { row: moveData.position.row, col: moveData.position.col };
                            if (validMoves.some(m => m.row === openingMove.row && m.col === openingMove.col)) {
                                return openingMove;
                            }
                        }
                    }
                }
            }

            // --- Phase 2: Positional Evaluation with Lookahead ---
            let bestMove = null;
            let bestScore = -Infinity;

            for (const move of validMoves) {
                // Start with the position's base value from the adapted matrix
                let score = positionValues[move.row][move.col];

                // Simulate the move using the API to observe its direct impact
                const result = api.simulateMove(board, player, move.row, move.col);

                if (result.valid) {
                    // Basic capture bonus
                    score += result.capturedCount * 5;

                    // Evaluate the resulting board state using the API's evaluator (reflects true rules)
                    const evaluationAfterMove = api.evaluateBoard(result.resultingBoard, player);

                    // Incorporate mobility score, adapted for potential reversed win condition
                    let mobilityContribution = evaluationAfterMove.mobilityScore * 10;
                    if (behaviors.winConditionReversedObserved) {
                        // If reversed, mobility for opponent might be good for us (fewer pieces)
                        // This is a complex adaptation; for now, maybe reduce mobility focus or even invert.
                        // A simpler adaptation: focus on reducing opponent's moves if reversed
                        mobilityContribution = -evaluationAfterMove.mobilityScore * 5; // Less mobility is good
                    }
                    score += mobilityContribution;

                    // Add a bonus for corners (if applicable, values from positionValues already handle this)
                    // (Corner values are usually high in positionValues, so direct addition might be redundant)

                    // Simple 1-ply minimax-like lookahead using API's evaluateBoard
                    // Simulate opponent's best response to this move
                    const opponent = (player === 1) ? 2 : 1;
                    const opponentValidMoves = api.getValidMoves(result.resultingBoard, opponent);

                    if (opponentValidMoves.length > 0) {
                        let worstCaseOpponentScore = Infinity; // Opponent tries to maximize their score

                        for (const oppMove of opponentValidMoves) {
                            const oppSimResult = api.simulateMove(result.resultingBoard, opponent, oppMove.row, oppMove.col);
                            if (oppSimResult.valid) {
                                const oppEvaluation = api.evaluateBoard(oppSimResult.resultingBoard, player).totalScore;
                                // We want to minimize opponent's gain, so we look for the worst-case for us
                                if (oppEvaluation < worstCaseOpponentScore) {
                                    worstCaseOpponentScore = oppEvaluation;
                                }
                            }
                        }
                        // Penalize our score by opponent's potential worst-case (for us) outcome
                        score += worstCaseOpponentScore * -0.5; // Multiply by a negative factor
                    } else {
                        // If opponent has no moves after our move, it's generally very good!
                        score += 100;
                    }


                    if (score > bestScore) {
                        bestScore = score;
                        bestMove = move;
                    }
                }
            }
            return bestMove;
        };
    }
}
\end{minted}

\section{Board Variation Types}
The Othello AI Arena introduces a diverse set of environmental variations to rigorously evaluate the adaptive and generalization capabilities of intelligent systems. These variations, which remain unseen to the AI until the analysis phase, probe different cognitive aspects crucial for artificial general intelligence. They are broadly categorized as follows:

\subsection{Structural Variations}
Structural variations modify the physical layout of the Othello board, challenging an AI's spatial reasoning, pathfinding, and dynamic re-evaluation of positional values.
\begin{itemize}
    \item \textbf{Board Size Alterations}: The game board deviates from the standard $8\times 8$ grid (Stage 1), including smaller (e.g., $6\times 6$, Stage 2) or larger dimensions. This tests an AI's ability to generalize its search strategies, positional evaluations, and heuristics to new scales.
    \item \textbf{Blocked Cells}: Impassable cells are introduced on the board, fundamentally altering connectivity and control areas. This necessitates dynamic re-evaluation of move validity, region control, and strategic pathfinding in a constrained space. An example of such a variation is the "$8\times 8$ (Partial C-Squares-cw)" stage (Stage 3), where specific cells are blocked.
\end{itemize}

\begin{figure}[h]
\centering
\begin{subfigure}[b]{0.3\textwidth}
    \centering
    \includegraphics[width=\textwidth]{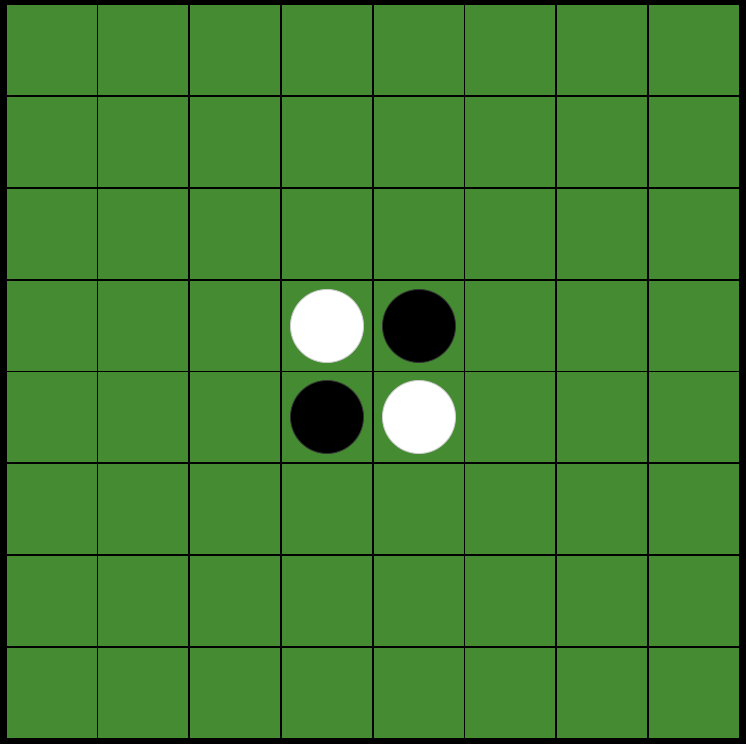}
    \caption{Standard $8\times 8$}
    \label{fig:standard_8x8}
\end{subfigure}
\hfill
\begin{subfigure}[b]{0.225\textwidth}
    \centering
    \includegraphics[width=\textwidth]{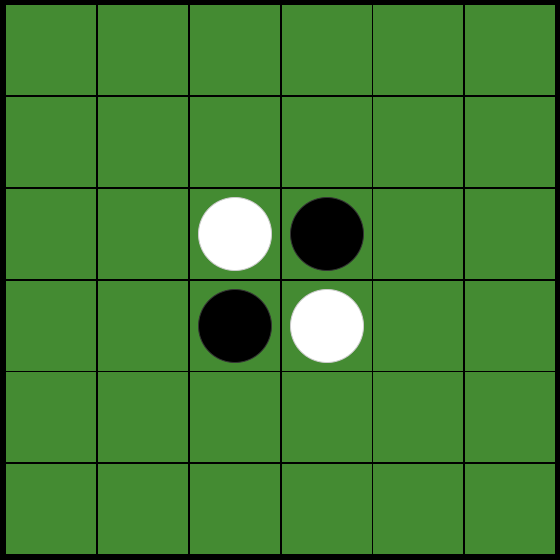}
    \caption{Small $6\times 6$}
    \label{fig:small_6x6}
\end{subfigure}
\hfill
\begin{subfigure}[b]{0.3\textwidth}
    \centering
    \includegraphics[width=\textwidth]{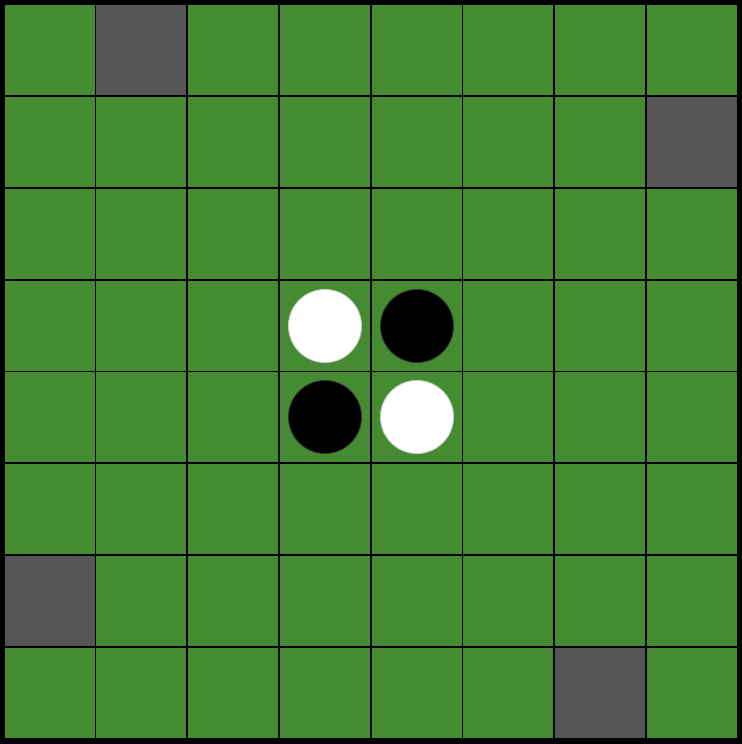}
    \caption{$8\times 8$ with blocked C-squares}
    \label{fig:blocked_cells_8x8}
\end{subfigure}
\caption{Basic spatial board variations with different board sizes and blocked cell configurations.}
\label{fig:basic_spatial_variations}
\end{figure}

Figure \ref{fig:basic_spatial_variations} illustrates some fundamental structural variations designed within our framework. Specifically, the "C-square" variation (Figure \ref{fig:blocked_cells_8x8}) introduces blocked cells that significantly impact movement patterns and strategic value maps.

\subsection{Rule Variations}
Rule variations fundamentally alter the game mechanics, demanding rapid rule induction and flexible adaptation of strategic thinking. These are particularly challenging as the AI must infer the very laws governing the environment.
\begin{itemize}
    \item \textbf{Capture Mechanics Modifications}: The core "sandwich" rule for capturing pieces can be altered. For instance, in an `ignore occlusion' variant, blocked cells do not halt capture lines, allowing pieces to be flipped over obstacles. This requires an AI to quickly update its move simulation and evaluation logic based on observed outcomes. A notable example is the "$8\times 8$ (C-Squares Occlusion Agnostic)" stage (Stage 12), which features this rule in our benchmark.
    \item \textbf{Turn Dynamics Alterations}: The rules governing whose turn it is can change. An example is the `fewer pieces continue' rule (Stage 13), where the player with fewer pieces takes consecutive turns, profoundly affecting temporal planning and requiring an understanding of game flow based on piece counts.
    \item \textbf{Winning Conditions Changes}: The objective of the game can be modified. For example, in "Reverse Othello," the player with the \emph{least} pieces at the end wins, directly challenging goal reorientation and counter-intuitive strategic thinking.
    \item \textbf{Move Restrictions}: Certain conditions might impose limitations on move validity, requiring dynamic re-evaluation of the action space.
\end{itemize}

\begin{figure}[h]
\centering
\begin{subfigure}[b]{0.43\textwidth}
    \centering
    \includegraphics[width=\textwidth]{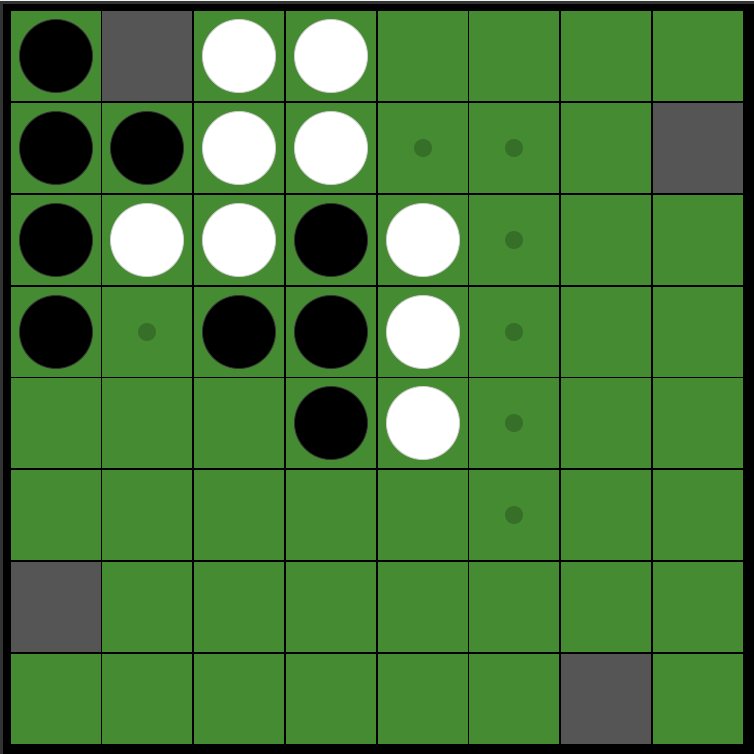}
    \caption{Standard Capture Mechanics (as typically found in Stage 3): For instance, placing a piece at e1 would not capture pieces beyond b1, as b1 is a blocked cell. This illustrates the typical logic where blocked cells halt capture lines.}
    \label{fig:standard_capture}
\end{subfigure}
\qquad
\begin{subfigure}[b]{0.43\textwidth}
    \centering
    \includegraphics[width=\textwidth]{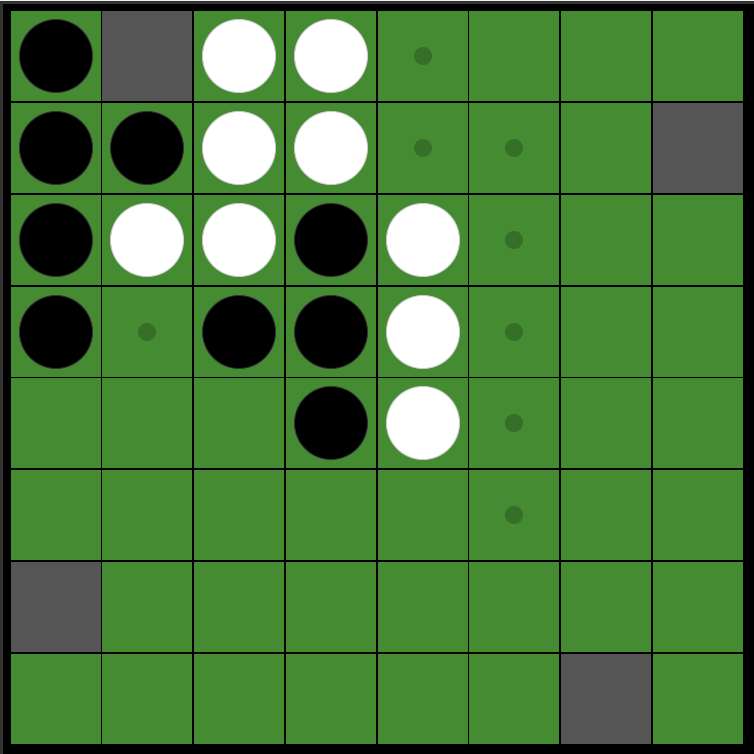}
    \caption{Occlusion-Agnostic Capture (Stage 12): Placing a piece at e1 \textit{can} capture pieces (like c1, d1) even across a blocked cell (b1). An intelligent system must infer this altered capture mechanic by observing the API's behavior.}
    \label{fig:occlusion_agnostic}
\end{subfigure}
\hfill
\caption{Advanced spatial and rule variations demonstrating complex environmental modifications. The AI must adapt its strategy based on inferred rule changes.}
\label{fig:advanced_variations}
\end{figure}

Figure \ref{fig:advanced_variations} visually demonstrates how a subtle rule change, like `ignore occlusion,' impacts gameplay and player perception. The intelligent system must discern this alteration through interaction with the API's \texttt{simulateMove} function.

\subsection{Initial State Variations}
These variations present non-standard starting configurations, which invalidate reliance on fixed opening strategies and demand dynamic early-game adaptation.
\begin{itemize}
    \item \textbf{Non-standard Initial Placement}: Pieces may begin in unusual positions, requiring the AI to dynamically generate early-game strategies instead of using predefined opening books.
    \item \textbf{Pre-placed Special Pieces}: (Potential future extension) Introducing pieces with unique properties or interactions from the outset.
    \item \textbf{Randomized Initialization}: Some initial board states might include random elements, adding to the novelty.
\end{itemize}

Table \ref{tab:variation_effects} summarizes the impact of each variation type on game dynamics and required strategic adaptations.

\begin{table}[h]
\caption{Impact of variation types on game dynamics and strategic adaptation.}
\label{tab:variation_effects}
\centering
\begin{tabular}{p{2.5cm}p{5cm}p{5cm}}
\toprule
\textbf{Variation Type} & \textbf{Impact on Game Dynamics} & \textbf{Strategic Implications} \\
\midrule
Structural Variations & Changes move patterns, accessibility of specific areas & Re-evaluation of positional values, shifts in area control strategies \\
\midrule
Rule Variations & Alters capture mechanisms, modifies turn dynamics & Changes risk-reward balance, requires adaptation of tempo strategies \\
\midrule
Initial State Variations & Modifies early game dynamics, affects opening patterns & Invalidates standard opening strategies, necessitates adaptive early-game play \\
\bottomrule
\end{tabular}
\end{table}

\section{Time Management Strategies in Intelligent System Development}

Participants develop intelligent systems for this benchmark must strategically allocate computational resources across three distinct temporal phases, each characterized by different constraints and optimization objectives.

\subsection{Leveraging $T_\text{development}$: The Extended Preparation Phase}

The development period $T_\text{development} \approx 30$ days provides participants with substantial computational budget to design and optimize their intelligent systems before deployment.

\paragraph{Foundation Building and API Mastery}
During $T_\text{development}$, participants must achieve comprehensive understanding of the arena architecture and API functionality through systematic experimentation on public stages $S_\text{public} = \{s_1, s_2, s_3\}$. The optimization objective is to develop an intelligent system $I$ such that the generated strategies $f_{s_i} = I(s_i, \text{API})$ outperform baseline strategies across all $s_i \in S_\text{public}$. This process enables participants to infer the benchmark's underlying design principles and adaptation requirements.

\paragraph{Generalization vs. Overfitting Trade-off}
While participants have access to $|S_\text{public}| = 3$ training stages and knowledge that evaluation will occur on $|S_\text{private}| = n$ hidden stages, they are encouraged to hypothesize potential configurations and construct augmented stages $S_\text{aug}$ for validation. However, the development approach must avoid creating systems that memorize solutions across $S_\text{public} \cup S_\text{aug}$, as this constitutes data interpolation rather than genuine adaptation capability. The fundamental challenge requires developing adaptation mechanisms that generalize effectively when $S_\text{train} \cap S_\text{test} = \emptyset$.

\paragraph{Temporal Resource Allocation Optimization}
A critical aspect of $T_\text{development}$ involves optimizing the allocation strategy for $T_\text{analysis}$. Let $\alpha, \beta, \gamma$ represent the proportion of analysis time dedicated to environment discovery, strategy adaptation, and parameter optimization respectively, where $\alpha + \beta + \gamma = 1$. An exemplary allocation might set $\alpha = 0.2$ for winning condition and dynamics identification, $\beta = 0.3$ for adapting existing high-performance strategies, and $\gamma = 0.5$ for self-play-based parameter refinement among top-performing agents. Since optimizing $(\alpha, \beta, \gamma)$ requires extensive experimentation and cannot be performed within the constrained $T_\text{analysis}$ window, these hyperparameters must be determined during $T_\text{development}$.

Similarly, participants must predetermine $T_\text{game}$ utilization strategies. Given that typical 8×8 Othello games require approximately 30 moves for completion, the challenge becomes distributing the total budget $T_\text{game} \approx 10$ seconds across individual move decisions $t_i$ such that $\sum_{i=1}^{30} t_i \leq T_\text{game}$ while maximizing strategic performance.

\subsection{Optimizing $T_\text{analysis}$: Rapid Environment Modeling}

The analysis phase $T_\text{analysis} \approx 60$ seconds requires efficient algorithms for environment discovery and strategy synthesis when presented with novel stage configuration $s \in S_\text{private}$.

\paragraph{Environment Dynamics Identification}
Upon encountering stage $s$ with unknown rule variations, the intelligent system must rapidly identify the winning condition $W(s)$ and transition dynamics $\mathcal{T}(s)$. This process leverages the provided API functions: \texttt{getValidMoves}$(b, p)$, \texttt{simulateMove}$(b, p, r, c)$, and \texttt{evaluateBoard}$(b, p)$, where $b$ represents board state, $p$ denotes player, and $(r,c)$ specifies position coordinates. The system must systematically explore the action space $\mathcal{A}(b, p)$ to construct an implicit model of environment dynamics within the temporal constraint.

\paragraph{Strategic Hypothesis Formation}
Following environment model construction, the system must synthesize strategy function $f_s: (\mathcal{B} \times \mathcal{P} \times \mathcal{A}) \rightarrow \mathcal{A}$ that maps board states, player identifiers, and valid action sets to selected actions. This synthesis process typically involves rapid simulation-based learning, where the system executes approximately 3000 self-play games using the API to gather statistical evidence for position evaluation and opening sequence optimization.

\subsection{Managing $T_\text{game}$: Constrained Real-Time Execution}

The game execution phase imposes the constraint $\sum_{\text{moves}} \text{Time}(f_s(b, p, M)) \leq T_\text{game} \approx 10$ seconds, requiring strategic time distribution analogous to bullet chess optimization~\cite{guga2022time}.

\paragraph{Theoretical Foundations}
World-championship Othello programs demonstrate efficient search budget allocation under extreme temporal constraints~\cite{othello1982world}. Progressive simulation techniques applied to perfect-information games optimize computational resources through adaptive budget adjustment~\cite{liao2023minizero}.

\paragraph{Optimal Time Allocation Strategies}
Empirically validated time management follows structured distribution patterns:
\begin{itemize}
\item \textbf{Phase-dependent Allocation:} Let $t_\text{opening}, t_\text{midgame}, t_\text{endgame}$ denote per-move time budgets for different game phases:
 \begin{itemize}
 \item Opening: $t_\text{opening} \in [0.5, 1.0]$ seconds for rapid development
 \item Midgame: $t_\text{midgame} \in [1.0, 2.0]$ seconds with higher allocation for tactical critical points
 \item Endgame: $t_\text{endgame} < 1.0$ second based on remaining search space
 \end{itemize}
\item \textbf{Premove Optimization:} Pre-computation of likely responses to minimize execution latency while maintaining adaptability to unexpected opponent moves
\item \textbf{Heuristic Approximation:} Replacement of computationally expensive search procedures with rapid pattern-based evaluation focusing on corner control and edge stability
\item \textbf{Memoization Strategies:} Caching of pre-computed positional value matrices and tactical sequences during $T_\text{analysis}$ for $O(1)$ lookup during $T_\text{game}$
\end{itemize}

\section{Game Logging and Data Format}
\label{app:logging}

The \thearena{} provides a comprehensive game logging system that records the details of each game played, including both standard matches and tournament games. This logging functionality is crucial for post-hoc analysis of adaptation strategies, debugging intelligent systems, and building valuable datasets for future research.

The platform allows users to save game logs in two primary formats: a human-readable text format and a machine-readable JSON format. This can be done via a "Save Log" function available in the user interface after a game or tournament is completed. These logs capture the full sequence of events and states, enabling participants and researchers to reconstruct games and analyze player behaviors and environmental interactions.

\subsection{Human-Readable Text Log}
\label{app:text_log}

The text log provides a simple, turn-by-turn record of the game in a format easily understandable by humans. It includes basic information about the game setup (strategies, stage) and the sequence of moves made by each player, along with the final score and winner. This format is particularly useful for quickly reviewing game flow and identifying specific sequences of moves.

A representative snippet of the human-readable text log format is shown below:

\begin{minted}[linenos, fontsize=\small, frame=lines, breaklines=true]{text}
=== Game 1 ===
Game started: Corners(B) vs Greedy(W) on Stage: Standard 8x8
Corners(B): d3
Greedy(W): c3
Corners(B): b3
Greedy(W): b2
Corners(B): b1
Greedy(W): e3
Corners(B): f3
Greedy(W): a1
Corners(B): c4
Greedy(W): g3
Corners(B): h3
Greedy(W): e2
Corners(B): d1
Greedy(W): a3
Corners(B): a2
Greedy(W): b4
Corners(B): a4
Greedy(W): a5
Corners(B): b5
Greedy(W): c5
Corners(B): a6
Greedy(W): c2
Corners(B): c1
Greedy(W): e1
Corners(B): f1
Greedy(W): d2
Corners(B): e6
Greedy(W): e7
Corners(B): f4
Greedy(W): g4
Corners(B): g2
Greedy(W): g1
Corners(B): h1
Greedy(W): f2
Corners(B): e8
Greedy(W): f5
Corners(B): g5
Greedy(W): h5
Corners(B): c6
Greedy(W): h2
Corners(B): h4
Greedy(W): b7
Corners(B): a8
Greedy(W): d7
Corners(B): h6
Greedy(W): h7
Corners(B): h8
Greedy(W): g6
Corners(B): c8
Greedy(W): f6
Corners(B): d6
Greedy(W): d8
Corners(B): f7
Greedy(W): f8
Corners(B): g8
Greedy(W): g7
Corners(B): c7
Greedy(W): b8
Corners(B): b6
Greedy(W): a7
Game over: Final score 27-37
White wins!
% ... (additional games if multiple are logged)
\end{minted}

\subsection{Machine-Readable JSON Log}
\label{app:json_log}

The JSON log provides a structured, machine-readable representation of the game data. This format is ideal for automated analysis, data processing, and integration with external tools for in-depth research. It contains detailed information for each game, including metadata, the initial board state, a sequence of moves with associated board states after each move, captured piece counts, time usage, and potentially analysis-specific data from intelligent systems.

A simplified structure and a representative snippet of the JSON log format are shown below. The full JSON file contains entries for each move, detailing the board state, player, position, captured pieces, and time taken for that move.

\begin{minted}[linenos, fontsize=\small, frame=lines, breaklines=true]{json}
[
  {
    "metadata": {
      "timestamp": "2025-05-16T07:04:50.912Z",
      "stageId": "stage-0",
      "stageName": "Standard 8x8",
      "blackStrategy": "Corners",
      "whiteStrategy": "Greedy",
      "blackScore": 27,
      "whiteScore": 37,
      "winner": 2, // 1 for Black, 2 for White, 0 for Tie
      "gameLength": 60 // Number of moves
    },
    "initialBoard": [
      // 2D array representing the board (0: Empty, 1: Black, 2: White, 3: Blocked)
      [0, 0, 0, 0, 0, 0, 0, 0],
      // ... more rows
      [0, 0, 0, 0, 0, 0, 0, 0]
    ],
    "moves": [
      {
        "player": 1, // 1 for Black, 2 for White
        "position": {"row": 2, "col": 3},
        "capturedCount": 1,
        "timeSpent": 127, // Time taken for this move in milliseconds (example field)
        "boardAfter": [
          // 2D array of the board state after this move
          [0, 0, 0, 0, 0, 0, 0, 0],
          // ... more rows
          [0, 0, 0, 0, 0, 0, 0, 0]
        ]
      },
      // ... additional move objects
    ],
    "analysisData": { // Optional: Data generated by the intelligent system during analysis
       "blackSystem": {
         "analysisTime": 58432, // Total analysis time in milliseconds
         "exploredPositions": 2743, // Example analysis metric
         // ... other analysis-specific data
       }
    }
  }
  // ... additional game objects if multiple games are logged
]
\end{minted}

The availability of both detailed, machine-readable JSON data and easily reviewable text logs significantly enhances the research utility and educational value of the \thearena{}. Researchers can use the JSON data for large-scale quantitative analysis and model training, while participants can use the text logs and visual replay (discussed in Section \ref{sub:architecture}) to understand game flow and debug their strategies.

\section{Human Adaptation to Novel Board Games: Observations}

To better understand the benchmark's relevance to human intelligence, I analyzed how humans adapt to novel board games using our time constraint framework, by observing an experienced board game player (one of our graduate student with chess ELO 1800+).

\subsection{Adaptation in Othello Arena Stages}

I examined adaptation patterns by having the participant navigate through progressively complex Othello Arena stages. Despite limited prior Othello experience beyond basic concepts like corner advantages, this provided an ideal case study for rapid environmental adaptation within our benchmark framework.

\paragraph{Stage 3 to Stage 12: Rule Variation Discovery}
The transition from Stage 3 (blocked cells with standard capture rules) to Stage 12 (identical board layout but modified capture mechanics allowing pieces to jump over obstacles) revealed sophisticated rule inference mechanisms. When initially attempting to play Stage 12 with Stage 3 strategies, the participant quickly recognized discrepancies between expected and actual game outcomes. Within two games, they formulated a crucial insight: "Stage 12's evaluation criteria are closer to Stage 1 (standard $8\times 8$) than Stage 3," demonstrating rapid similarity matching between environmental variants.

The participant's rule discovery process followed a systematic three-step cognitive pattern. First, they conducted internal simulations, stating "If I place here, the opponent will likely respond there," followed by intuitive strategic evaluation based on domain knowledge accumulated from previous games. Second, they analyzed opponent moves to infer underlying strategic principles, noting "I try to understand what the opponent wants to achieve by observing their moves." Third, they relied on visual similarity assessment, explaining that board layouts that appear similar often share similar rule structures, allowing for rapid strategic transfer.

\paragraph{Domain Knowledge Transfer and Linguistic Abstraction}
Throughout the adaptation process, the participant leveraged abstract strategic principles commonly used in perfect-information games. They employed heuristic guidelines such as "occupy the center first" and "secure corners early," demonstrating how linguistic abstractions facilitate rapid strategy transfer across game variants. Remarkably, the participant noted that when teaching complete novices, strategic concepts are often communicated through everyday language expressions like "good shape" or "strong position," suggesting that natural language serves as a bridge between complex strategic reasoning and rapid knowledge transfer.

When encountering rule ambiguity, the participant explicitly articulated their uncertainty management strategy: "In situations where I'm unsure about the rules, I tend to adopt more conservative, broadly applicable strategies rather than specific optimizations. If I know I have multiple attempts, I'll use the first game for exploration—trying various moves to understand the dynamics—while in subsequent games I can optimize based on learned patterns."

\paragraph{Stage 13: Turn Dynamics and Hypothesis Refinement}
Stage 13 introduced a novel turn mechanism where players with fewer pieces on the board could take consecutive turns. This created a more complex adaptation challenge, as the participant initially formed an incorrect hypothesis about the underlying mechanics. They interpreted the rule as "players who move quickly can take multiple consecutive turns," leading to a time-pressure based strategy rather than the actual piece-count based mechanic.

Despite the misinterpretation, the participant demonstrated sophisticated hypothesis testing behavior. They noted, "When I suspect the rules have changed, I engage in more exploratory play during the first game, deliberately trying various moves to understand the new dynamics. This is different from optimization-focused play where I'm trying to win—here I'm trying to learn." The participant's approach shifted toward broader, more generalizable strategies when facing uncertainty, prioritizing information gathering over immediate tactical advantage.

\paragraph{Cognitive Mechanisms and Strategic Flexibility}
The adaptation process revealed several key cognitive mechanisms that distinguish human learning from current AI approaches. The participant demonstrated within-episode learning, continuously refining their environmental model during gameplay rather than requiring separate training phases. They exhibited rapid transfer learning, immediately drawing analogies to familiar game patterns and applying relevant strategic principles. Most notably, they showed meta-cognitive awareness, explicitly reasoning about their own learning process and adjusting their exploration-exploitation balance based on the perceived novelty of the environment.

When asked about their approach to reward function estimation in novel environments, the participant explained: "I rely heavily on domain knowledge from previous games. When I encounter a new situation, I first look for similar patterns I've seen before, then gradually adjust my evaluation based on what works and what doesn't. The key is having these broad strategic principles that work across different games, then fine-tuning them to the specific rules I discover."

This observation suggests that effective adaptation in complex environments may require both rapid pattern recognition capabilities and flexible strategic frameworks that can be quickly reconfigured based on environmental feedback—characteristics that current AI systems struggle to replicate efficiently.

\subsection{Adaptation in YINSH Game}

As he demonstrated the ability to defeat intermediate AI opponents and human players in both Othello Arena's hidden stages and Ataxx within 2--3 games. To explore the limits of rapid human adaptation, I introduced him to YINSH\footnote{\url{www.yinsh.net}}, more complex abstract strategy game that combines elements of Othello and Gomoku with additional strategic layers involving ring management and multi-objective optimization.

\paragraph{Rapid Environment Modeling and Rule Discovery}
During the first game (\string~30 moves, \string~15 minutes), the player demonstrated remarkable efficiency in discovering environment dynamics within a single episode. They correctly identified the core winning condition (graduating three rings by forming consecutive lines of five pieces), understood the dual mechanics of piece placement and ring movement, and began forming hypotheses about capture mechanisms. Notably, approximately 70\% of the game's fundamental mechanics were grasped within this first encounter. However, some misconceptions emerged initially—for instance, believing that forming exactly five pieces (rather than five or more) was required for ring graduation, and uncertainty about which ring to remove when multiple options were available. This rapid yet imperfect rule acquisition represents an extremely compressed learning timeline compared to traditional reinforcement learning approaches, which typically require hundreds or thousands of episodes for comparable environmental understanding.

\paragraph{Strategic Hypothesis Formation and Transfer Learning}
The second game revealed sophisticated strategy development through active hypothesis testing and knowledge transfer. The player immediately drew analogies to familiar games, recognizing blocking and connection patterns similar to Go and Gomoku. They applied prior strategic knowledge about piece activity, stating that "in perfect information games, it's important to keep all pieces active" and began implementing multi-step planning with explicit discussion of opponent modeling ("they're attacking, so I need to defend" and "this move is safe for now"). The player exhibited within-episode learning, continuously refining their understanding of positional value, ring positioning, and tactical combinations as the game progressed. This demonstrates the human capacity for rapid transfer learning and strategic reasoning under minimal data exposure—a stark contrast to typical online reinforcement learning which requires numerous episodes for comparable strategic development.

\paragraph{Time Constraint Analysis}
Interpreting human adaptation through our framework reveals interesting parallels and contrasts with AI systems. Human learning effectively combines $T_\text{analysis}$ and $T_\text{game}$ in a fluid, interleaved manner similar to ARC-PRIZE's structure, where each 15-minute game serves simultaneously as analysis time (discovering rules and forming strategies) and execution time (implementing tactical decisions). The key insight is that humans achieve intermediate-level play within merely 2--3 attempts (30--45 minutes total), while maintaining the ability to deliberate and refine hypotheses in real-time during gameplay. Unlike our benchmark's clear separation between analysis and execution phases, humans seamlessly transition between environment modeling, strategy formation, and tactical execution within the same episode.

\paragraph{Implications for AGI Development}
This observation pattern aligns with recent developments in test-time compute scaling, where systems like GRPO~\cite{deepseek2024} and test-time reinforcement learning~\cite{snell2024scaling} demonstrate that complex novel tasks require active hypothesis building and refinement during inference, rather than relying solely on pre-trained knowledge. The human adaptation pattern suggests that effective artificial general intelligence may need similar capabilities for dynamic reasoning and rapid environmental modeling under minimal data exposure. Furthermore, the stark efficiency gap between human adaptation (2--3 games) and current AI approaches (typically requiring thousands of simulations as seen in our template system) highlights a fundamental challenge: while our benchmark systems can eventually match or exceed human performance through extensive simulation within $T_\text{analysis}$, they lack the elegant efficiency of human learning that combines rapid rule discovery, strategic transfer, and real-time adaptation within a unified cognitive process.

\end{document}